\documentclass[10pt,journal,compsoc]{IEEEtran}

\usepackage[dvipsnames,svgnames]{xcolor}

\ifCLASSOPTIONcompsoc
  \usepackage{cite}
\else
  \usepackage{cite}
\fi

\ifCLASSINFOpdf
  \usepackage[pdftex]{graphicx}
\else
  \usepackage[dvips]{graphicx}
\fi

\usepackage{amsmath}
\usepackage{amssymb}
\usepackage{amsfonts} 
\usepackage{pax}

\usepackage{hyperref}
\usepackage{url}
\usepackage{color}
\usepackage{colortbl}
\usepackage{soul}
\usepackage{booktabs} %
\usepackage{multirow}
\usepackage{subcaption}
\usepackage[ruled,vlined,linesnumbered]{algorithm2e}

\usepackage{rotating,siunitx} %
\usepackage{tabularx}
\usepackage{hyphenat}
\usepackage{longtable}
\usepackage{nicefrac}       
\usepackage{microtype}      

\usepackage{float}

\usepackage{etoolbox}

\usepackage{setspace}

\makeatletter
\patchcmd{\@makecaption}
  {\scshape}
  {}
  {}
  {}
\makeatother

\usepackage[noend]{algpseudocode} 

\usepackage{pifont}
\newcommand{\xmark}{\ding{55}}%

\newcommand{\etal}{\textit{et al}.}
\newcommand{\ie}{\textit{i}.\textit{e}.}
\newcommand{\eg}{\textit{e}.\textit{g}.}
\definecolor{commentGreen}{rgb}{0,0.5,0.05}

\definecolor{LightRed}{rgb}{1,0.92,0.92}
\definecolor{LightOrange}{rgb}{1,0.95,0.88}
\definecolor{LightYellow}{rgb}{1.0,1.0,0.8}
\definecolor{LightGreen}{rgb}{0.9,1.0,0.88}
\definecolor{LightCyan}{rgb}{0.8,1,1}
\definecolor{LightBlue}{rgb}{0.9,0.94,1}
\definecolor{LightIndigo}{rgb}{0.92,0.9,1}
\definecolor{LightMagenta}{rgb}{0.96,0.86,1}
\definecolor{DirtyWhite}{rgb}{0.96,0.96,0.96}

\soulregister\ref7
\soulregister\cite7
\soulregister\etal7

\begin{document}
%
\title{Curriculum-DPO++: Direct Preference Optimization via Data and Model Curricula for Text-to-Image Generation}

\author{{Florinel-Alin}~Croitoru,
        {Vlad}~Hondru,
        {Radu Tudor}~Ionescu,~\IEEEmembership{Member,~IEEE,}
        {Nicu}~Sebe,~\IEEEmembership{Senior Member,~IEEE,}
        and~Mubarak~Shah,~\IEEEmembership{Fellow,~IEEE}
\IEEEcompsocitemizethanks{\IEEEcompsocthanksitem F.A. Croitoru, V. Hondru and R.T. Ionescu are with the Department of Computer Science, University of Bucharest, Bucharest, Romania. F.A. Croitoru and V. Hondru have contributed equally. R.T. Ionescu is the corresponding author.\protect\\
E-mail: raducu.ionescu@gmail.com
\IEEEcompsocthanksitem N. Sebe is with the University of Trento, Italy.
\IEEEcompsocthanksitem M. Shah is with the Center for Research in Computer Vision (CRCV), University of Central Florida, Orlando, FL, US.}
\thanks{Manuscript received February 10, 2026; revised June 27, 2026.}}

%
%

\markboth{IEEE Transactions on Pattern Analysis and Machine Intelligence,~Vol.~55, No.~1, August~2026}%
{Croitoru \MakeLowercase{\textit{et al.}}: Data and Model Curriculum Direct Preference Optimization for Image Generation}


\IEEEtitleabstractindextext{%
\begin{abstract}
Direct Preference Optimization (DPO) has been proposed as an effective and efficient alternative to reinforcement learning from human feedback (RLHF). However, neither RLHF nor DPO take into account the fact that learning certain preferences is more difficult than learning other preferences, rendering the optimization process suboptimal. To address this gap in text-to-image generation, we recently proposed Curriculum-DPO, a method that organizes image pairs by difficulty, ensuring that increasingly challenging examples are progressively sampled and presented to a generative (diffusion or consistency) model. In this paper, we introduce Curriculum-DPO++, an enhanced method that combines the original data-level curriculum with a novel model-level curriculum. More precisely, we propose to dynamically increase the learning capacity of the denoising network as training advances. We implement this capacity increase via two mechanisms. First, we initialize the model with only a subset of the trainable layers used in the original Curriculum-DPO. As training progresses, we sequentially unfreeze layers until the configuration matches the full baseline architecture. Second, as the fine-tuning is based on Low-Rank Adaptation (LoRA), we implement a progressive schedule for the dimension of the low-rank matrices. Instead of maintaining a fixed capacity, we initialize the low-rank matrices with a dimension significantly smaller than that of the baseline. As training proceeds, we incrementally increase their rank, allowing the capacity to grow until it converges to the same rank value as in Curriculum-DPO. Furthermore, we propose an alternative ranking strategy to the one employed by Curriculum-DPO. While the original approach relies on external reward models to rank samples, we introduce an implicit scoring mechanism derived from prompt perturbations. This approach is particularly useful when auxiliary reward models are not available. Finally, we compare Curriculum-DPO++ against Curriculum-DPO and other state-of-the-art preference optimization approaches on nine benchmarks, outperforming the competing methods in terms of text alignment, aesthetics and human preference. Our code is available at \url{https://github.com/CroitoruAlin/Curriculum-DPO}.
\end{abstract}
}

\maketitle

\setlength{\abovedisplayskip}{3.5pt}
\setlength{\belowdisplayskip}{3.5pt}

\IEEEdisplaynontitleabstractindextext

\IEEEpeerreviewmaketitle

\IEEEraisesectionheading{\section{Introduction}\label{sec:introduction}}
\IEEEPARstart{D}{iffusion} models \cite{Croitoru-TPAMI-2023,ho-NeurIPS-2020,sohl-icml-2015,song-NeurIPS-2019} represent a family of generative models that gained significant traction in image generation tasks, largely due to their impressive generative capabilities. One of the tasks where these models excel is text-to-image generation \cite{avrahami-CVPR-2022,gu-CVPR-2022,rombach-CVPR-2022,Saharia-NeurIPS-2022}, as they are capable of generating images that are both aesthetic and well aligned with the input text (prompt). However, state-of-the-art diffusion models that have been widely adopted by the community, e.g.~Stable Diffusion \cite{rombach-CVPR-2022}, are typically heavy in terms of training (and even inference) time. To this end, several studies \cite{Black-ICLR-2024,Fan-NeurIPS-2023,Luo-Arxiv-2023b,Wallace-arxiv-2023} proposed novel training methods to efficiently fine-tune pre-trained diffusion models. Some of these training frameworks \cite{Black-ICLR-2024,Luo-Arxiv-2023b,Wallace-arxiv-2023} were originally introduced for Large Language Models (LLMs) \cite{Christiano-NeurIPS-2017,Hu-ICLR-2022,Rafailov-NeurIPS-2023}, another family of models that are notoriously hard to train on a few GPUs \cite{Schwartz-CACM-2020,Strubell-ACL-2019}. This is also the case of Direct Preference Optimization (DPO) \cite{Rafailov-NeurIPS-2023}, a method used to fine-tune LLMs, which was originally proposed as an effective and efficient alternative to reinforcement learning from human feedback (RLHF) \cite{Christiano-NeurIPS-2017}. DPO bypasses the need to fit a reward model by harnessing a mapping between reward functions and optimal policies, which enables the direct optimization of the LLM to adhere to human preferences. DPO was later extended to diffusion models \cite{Wallace-arxiv-2023}, showcasing similar benefits in image generation. Despite the significant benefits brought by Diffusion-DPO \cite{Wallace-arxiv-2023} and related methods \cite{Black-ICLR-2024,Fan-NeurIPS-2023,Luo-Arxiv-2023b}, there are still observable gaps in terms of various factors, such as text alignment, aesthetics and human preference (see Figures \ref{qualitative_text_align},  \ref{qualitative_human_preference} and \ref{qualitative_aesthetics}). Therefore, more advanced adaptation techniques are required to reduce these gaps. 

\begin{figure*}[t]
 \begin{center}
  \centerline{\includegraphics[width=0.9\textwidth]{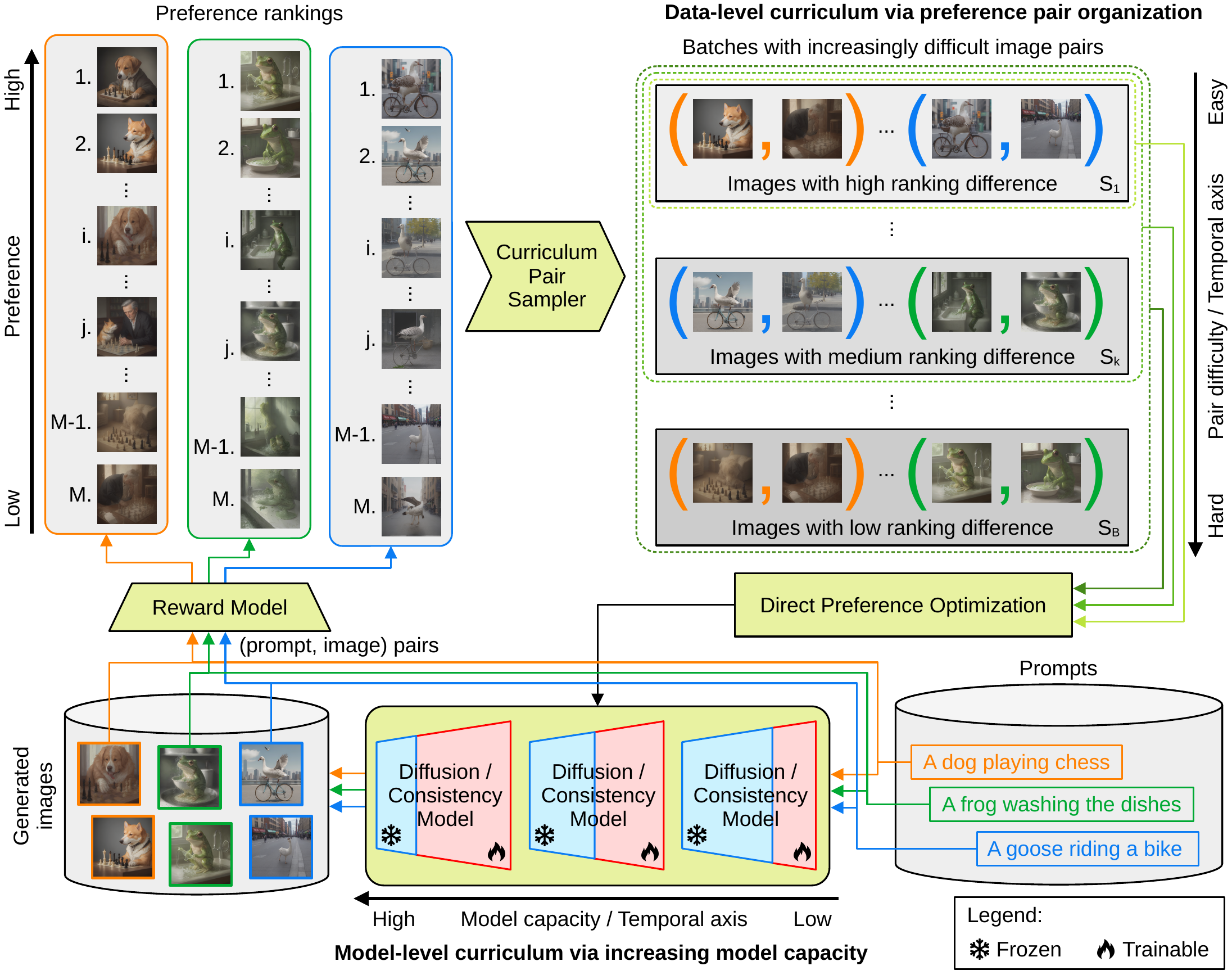}}
  \vspace{-0.22cm}
  \caption{An overview of Curriculum-DPO++. Images generated by a diffusion / consistency model and their prompts are passed through a reward model, obtaining a preference ranking for each prompt. Next, image pairs of various difficulty levels are generated and organized into batches, such that the initial batch contains easy pairs (with high difference in terms of preference scores) and subsequent batches contain increasingly difficult pairs (the difference in terms of preference is gradually decreased). The diffusion / consistency model is trained via Direct Preference Optimization (DPO) based on easy-to-hard data-level curriculum. Curriculum-DPO++ gracefully combines data-level and model-level curricula, increasing the learning capacity of the model to accommodate for the more complex examples gradually introduced during training. The learning capacity is expanded by unfreezing neural layers and increasing the rank of LoRA matrices. Best viewed in color.}
  \label{fig_pipeline}
  \vspace{-0.55cm}
\end{center}
\end{figure*}

One of the advancements that boosted the performance of Diffusion-DPO is our recent work, Curriculum-DPO~\cite{Croitoru-CVPR-2025}, which was published at CVPR 2025. Curriculum-DPO stems from the fact that the preference pairs in Diffusion-DPO are randomly sampled during training, which is suboptimal. To mitigate this issue, Curriculum-DPO organizes the pairs according to their complexity with respect to the preference learning task, from easy to hard. The method comprises two stages. In the first stage, we employ a reward model to rank the images generated for each prompt according to their preference score. In the second stage, pairs of samples with different difficulty levels are created, leveraging the rank difference between samples as a measure of difficulty. Samples that are far apart in the preference ranking are considered to form easy pairs. In contrast, samples that are close in the ranking form hard pairs, since choosing the preferred sample is less obvious. The sampled pairs are split into batches according to their difficulty level, allowing to organize the batches in a meaningful order, from easy to hard. This essentially generates an optimization process based on curriculum learning \cite{Bengio-ICML-2009}. However, in the early training stages, a high capacity model may be prone to overfitting on easy examples, hurting performance and generalization. In this work, we propose to enhance Curriculum-DPO by synchronizing the original data-level curriculum with a novel model-level curriculum, as illustrated in Figure \ref{fig_pipeline}. We implement the model-level curriculum via two complementary mechanisms. The first mechanism guides the progressive expansion of trainable layers. We initialize the training with a minimal subset of neural components, comprising only the bottleneck and the innermost downsampling and upsampling blocks. As we update the set of pairs in the data-level curriculum, we jointly expand the set of trainable layers by including additional outer downsampling and upsampling blocks. The second mechanism to realize the model-level curriculum operates on Low-Rank Adaptation (LoRA) matrices, since we run our fine-tuning using this procedure. We start with a small dimension for the low-rank matrices and, as we progress with the data-level curriculum, we also increase the dimension of the matrices. When we perform this dimension update, we copy the previously learned weights into the top-left part of the new larger matrices. The data-level and model-level curricula are gracefully integrated to adjust the learning capacity of the model to the current complexity of the training examples, resulting in a new framework, called Curriculum-DPO++.

As a secondary contribution, we introduce a reward-free alternative to the initial ranking stage of Curriculum-DPO. Rather than relying on an external reward model, we construct synthetic preference pairs and derive their difficulty rankings directly through prompt embedding perturbations. Specifically, we apply varying degrees of masking to the text embeddings. The winning sample is universally defined as the image generated with a masking ratio of 0 (the clean prompt). The losing sample corresponds to the perturbed prompt. We then map the masking ratio to the curriculum difficulty. The pairs with a smaller masking ratio are regarded as difficult (since the degradation is subtle), while those with a higher masking ratio are considered easy (as the heavy masking renders the degradation obvious).

We carry out experiments across nine evaluation benchmarks to assess the effectiveness of our Curriculum-DPO enhancements in adapting text-to-image generative models to different reward models. We first tackle the text alignment task, employing the Sentence-BERT \cite{reimers-EMNLP-2019} similarity metric to evaluate the correspondence between the actual text prompt and the caption provided by the LLaVA model \cite{liu-NeurIPS-2024} for each generated image. Next, we focus on enhancing the visual aesthetics of the images, where we use the LAION Aesthetics Predictor \cite{Schuhmann-laion-2022} as the reward model. Finally, for the third task, we increase the human preference rate by employing HPSv2 \cite{Wu-arXiv-2023} as the reward model. Each of the three tasks is studied on three distinct datasets. The quantitative and qualitative results on all datasets show that Curriculum-DPO++ consistently outperforms the original Curriculum-DPO and other state-of-the-art fine-tuning strategies, such as Diffusion-DPO \cite{Rafailov-NeurIPS-2023, Wallace-arxiv-2023} and Denoising Diffusion Policy Optimization (DDPO) \cite{Black-ICLR-2024}. 


In summary, our contribution is fivefold:
\begin{itemize}
    \item 
    We provide an extended presentation of Curriculum-DPO \cite{Croitoru-CVPR-2025}, a training regime for diffusion and consistency models, which enhances DPO via data-level curriculum learning.
    \item 
    We provide an extended presentation of our DPO adaptation for consistency models, termed Consistency-DPO \cite{Croitoru-CVPR-2025}.
    \item 
    We introduce a novel model-level curriculum framework that enhances Curriculum-DPO by increasing its learning capacity along with the complexity of the preference pairs, resulting in a new framework called Curriculum-DPO++.
    \item 
    We propose a reward-free alternative to Curriculum-DPO.
    \item 
    We demonstrate the superiority of Curriculum-DPO++ over Curriculum-DPO and other state-of-the-art training alternatives on nine evaluation benchmarks.
\end{itemize}

\vspace{-0.2cm}
\section{Related Work}
\label{sec: related_work}

\textbf{Diffusion models.}
In recent years, diffusion models have gained a lot of traction \cite{Croitoru-TPAMI-2023, ho-NeurIPS-2020, rombach-CVPR-2022, sohl-icml-2015, song-NeurIPS-2019, song-ICLR-2021}, becoming one of the state-of-the-art generative methods, due to their capability to synthesize diverse images with high-fidelity, greatly surpassing Generative Adversarial Networks (GANs) \cite{dhariwal-NeurIPS-2021}, the prior leading approach. Given their wide adoption, significant efforts were undertaken to improve the capability of diffusion models across multiple dimensions. One dimension regards the introduction of conditioning mechanisms for different modalities, \eg~text prompts \cite{avrahami-CVPR-2022,gu-CVPR-2022,ramesh-arXiv-2022,rombach-arXiv-2022}, images \cite{baranchuk-arXiv-2021, meng-arXiv-2021, saharia-SIGGRAPH-2022} or class labels \cite{chao-ICLR-2022, ho-arXiv-2021, salimans-arXiv-2022}. Another dimension is finding new downstream applications, \eg~inpainting \cite{lugmayr-CVPR-2022, nichol-ICML-2022}, super-resolution \cite{daniels-NeurIPS-2021, saharia-TPAMI-2022}, medical imaging \cite{chung-MIA-2022, wyatt-CVPR-2022}, etc. Despite the important research endeavors of both academia and industry, the main limitation of diffusion models is represented by their slow inference speed. More precisely, diffusion models need to execute many sequential denoising operations, which heavily slows down the sampling time. Consequently, considerable attention has been directed towards the sampling process through improved scheduling strategies \cite{liu-ICLR-2022, nichol-ICML-2021, song-ICLR-2021b}. Building towards the goal of reducing inference time, Song \etal~\cite{Song-ICML-2023} proposed consistency models, a new family of diffusion models requiring less denoising steps. Consistency models were further developed in follow-up studies, \eg~\cite{Luo-arXiv-2023, Song-ICLR-2024}. To the best of our knowledge, DPO was not studied in conjunction with consistency models before our preliminary work \cite{Croitoru-CVPR-2025}. In this work, we describe in detail how DPO can be adapted to consistency models.

\noindent
\textbf{Controllable generation for diffusion models.}
Guiding the image generation process in diffusion models using various signals has been a major research focus. The initial approaches leveraged the gradients of a classifier to steer synthesis \cite{chao-ICLR-2022, dhariwal-NeurIPS-2021, song-ICLR-2021}. Ho \etal~\cite{ho-NeurIPS-2021} later refined by introducing classifier-free guidance for conditional generation.

Building upon the same line of work, Zhang \etal~\cite{Zhang-ICCV-2023} proposed ControlNet, a framework that enables an additional control input (mostly for the structure) to any text-to-image diffusion model, without retraining the original model. Their method involves several architectural modifications. While keeping the original U-Net intact, trainable copies of the encoder blocks and middle blocks are created for the control input. These are further integrated with the original decoder blocks with zero-convolutional layers.

Due to their great success, the recent advances in LLMs have also been quickly adopted in the development of diffusion models. Hu \etal~\cite{Hu-ICLR-2022} introduced Low-Rank Adaptation (LoRA), a parameter-efficient fine-tuning methodology for LLMs through low-rank matrix factorization of selected transformer layers, while preserving the pre-trained parameters. LoRA brings improved performance with reduced memory overhead. Luo \etal~\cite{Luo-Arxiv-2023b} employed LoRA \cite{Hu-ICLR-2022} on consistency models, demonstrating better results, while requiring less memory.

A fundamental limitation of standard LLM training objectives is that they fail to adequately capture human preferences, while commonly-used metrics (\eg~BLEU \cite{papineni-ACL-2002}) are imperfect proxies. This motivated the switch from standard supervised training to reinforcement learning \cite{bai-arXiv-2022, havrilla-arXiv-2024, Ziegler-arXiv-2020}. Lee \etal~\cite{lee-ArXiv-2023} demonstrated improved text-image alignment by training a reward model on data labeled from human feedback. The reward model is further used to guide the fine-tuning of a text-to-image diffusion model. Policy gradients methods, instrumental for the success of LLMs, have also been integrated into diffusion models. While Fan \etal~\cite{Fan-NeurIPS-2023} demonstrated that such an algorithm leads to an improved text-image alignment, Black \etal~\cite{Black-ICLR-2024} applied their preference optimization method to enhance multiple aspects of synthesis: compressibility, aesthetics, as well as prompt alignment. More recently, Rafailov \etal~\cite{Rafailov-NeurIPS-2023} introduced Direct Preference Optimization (DPO), which eliminates the intermediate reward model and enables fine-tuning LLMs directly on the preference, by providing a pair of winning and losing samples. DPO demonstrates superior performance than several Proximal Policy Optimization methods \cite{ouyang-NeurIPS-2022, schulman-arXiv-2017, Ziegler-arXiv-2020}. Wallace \etal~\cite{Wallace-arxiv-2023} integrated DPO into diffusion models through a custom objective function and training procedure, reporting improvements on subjective metrics, such as visual aesthetics or prompt alignment. To the best of our knowledge, curriculum learning was first applied in conjunction with controllable generation methods for diffusion models in our preliminary study \cite{Croitoru-CVPR-2025}. In this study, we introduce an enhanced version of Curriculum-DPO \cite{Croitoru-CVPR-2025}, which integrates data and model curricula in a meaningful way.

\noindent
\textbf{Curriculum learning.} Inspired by how people learn, Bengio \etal~\cite{Bengio-ICML-2009} proposed \emph{curriculum learning}, a learning paradigm to train neural networks starting with an easier task, and then progressively increasing its difficulty. Despite the early introduction of curriculum learning more than a decade ago, it remains an active research direction \cite{Soviany-IJCV-2022,Wang-ICCV-2023,Wang-ACMMM-2024}, usually being integrated into the learning process of numerous neural models. Soviany \etal~\cite{Soviany-IJCV-2022} established four categories of curriculum learning, according to the level of operation: data, model, task, and optimization objective. Among these, data-level and model-level curricula are more popular, essentially because of their wider applicability across tasks. Data-level curriculum studies how organizing samples based on their complexity influences the training process \cite{Bengio-ICML-2009, Ionescu-CVPR-2016,jiang-AAAI-2015, kuman-NeurIPS-2010, Lao-ACMMM-2021,Soviany-CVIU-2021,Yu-ICCV-2025}. Model-level curriculum focuses on increasing the learning power of the model during training \cite{karras-ICLR-2018, Madan-WACV-2024, morerio-ICCV-2017, Sinha-NIPS-2020, Croitoru-IJCV-2025}, essentially eliminating the need to assess sample difficulty. A wide number of works in computer vision applied various forms curriculum learning \cite{Soviany-IJCV-2022, Wang-ICCV-2023}. Nevertheless, its application in image generation has been rather limited, being mainly focused on GANs \cite{doan-AAAI-2019, ghasedi-CVPR-2019, karras-ICLR-2018, soviany-wacv-2020}. Given the recent surge of interest in diffusion models, easy-to-hard curriculum learning has been explored for this class of generative models \cite{kim-arXiv-2024,xu-arXiv-2024, na-CVPR-2025}. Kim \etal~\cite{kim-arXiv-2024} and Xu \etal~\cite{xu-arXiv-2024} advocated to start the training from higher diffusion timesteps (with more noise) and progress to lower timesteps (with less noise). Their methods are completely different from our work, as we concentrate on the difficulty with respect to the preference scores rather than the noise levels. Building upon our previous work \cite{Croitoru-CVPR-2025}, Na \etal~\cite{na-CVPR-2025} proposed a fine-tuning strategy based on Curriculum-DPO for diffusion models. While very similar to our method, the main distinction is that their winning samples are based on real images, except for the first stage. Moreover, their training stages (easy, intermediate and hard) are related to the provenance of the winning samples. To the best of our knowledge, we are the first to combine data-level and model-level curricula by adjusting the learning capacity of the model with respect to the difficulty of the training samples. As the difficulty grows from easy to hard, the learning capacity is proportionally increased, avoiding overfitting on easy examples in the early training stages.

\vspace{-0.2cm}
\section{Method}
\label{sec: method}

Before diving into the details behind Curriculum-DPO++, we briefly discuss preliminaries regarding diffusion models, consistency models, DPO, and Diffusion-DPO. For a more detailed introduction of these concepts, please refer to Appendix \ref{supp_prelim}.

\vspace{-0.2cm}
\subsection{Preliminaries}
\vspace{-0.1cm}
\textbf{Diffusion models.} Diffusion models are probabilistic generative models that reconstruct data samples, $x_0 \sim p(x_0)$, from Gaussian noise. The training consists of a forward diffusion process that perturbs each data sample into a noisy version $x_t$ via a transition $x_t = \alpha_t x_0 + \sigma_t \epsilon$, where $\epsilon \sim \mathcal{N}(0, \mathbf{I})$ and $\{\alpha_t, \sigma_t\}_{t=1}^T$ defines the variance schedule. Learning is formulated as a noise-prediction task. We optimize a neural network $\epsilon_\theta(x_t, t)$, with parameters $\theta$, to approximate the underlying noise component $\epsilon$ using the standard mean squared error loss:
\begin{equation}
    \!\!\mathcal{L}_{\text{simple}}\!=\! \mathbb{E}_{t \sim \mathcal{U}(1, T), \epsilon \sim \mathcal{N}(0, \mathbf{I}), x_0 \sim p(x_0)} \!\left\lVert \epsilon_t\!-\!\epsilon_\theta(x_t,t)\right\rVert^{2}_2\!.
\end{equation}
Sampling is achieved by reversing this trajectory. We start with pure noise $x_T \sim \mathcal{N}(0, \mathbf{I})$ and iteratively refine the image using $\epsilon_\theta$. At each step $t$, we subtract the predicted noise from the image, pushing the image towards the original data manifold.

\noindent
\textbf{Consistency models.} Consistency models \cite{Song-ICML-2023, Luo-arXiv-2023, Song-ICLR-2024} leverage the deterministic Probability Flow ODE (PF-ODE) formulation of diffusion \cite{song-ICLR-2021}. They are trained to map arbitrary states $x_t$ on the solution trajectory directly to the origin $x_0$. This property is induced via a self-consistency constraint, requiring the model output to be identical for two points belonging to the same ODE trajectory. The training objective minimizes the distance between the current model output and a target estimated from the previous step:
\begin{equation}
\label{eq_consistency_distillation}
    \mathcal{L}_{\text{CD}}(\phi) = d(f_\phi(x_{t_{n+1}}, t_{n+1}), f_{\phi^{-}}(\hat{x}_{t_n}^{\theta}, t_n)),
\end{equation}
where $\hat{x}_{t_n}^\theta$ is a one-step estimate of the next state derived by a pre-trained frozen diffusion model, $\epsilon_\theta$, using a numerical ODE solver. The target network utilizes exponential moving average (EMA) weights $\phi^{-}$ to stabilize training, and $N$ represents the discretization of the time interval $[0, T]$, $n \sim \mathcal{U}(1, N)$, $\phi$ are the trainable parameters of the consistency model, and $d$ is a distance metric.

\noindent
\textbf{Direct Preference Optimization (DPO).} 
To align a generative model with human preferences, Rafailov~\etal~\cite{Rafailov-NeurIPS-2023} introduced DPO. This approach fine-tunes models directly on preference pairs $(x_0^w, x_0^l, c)$, where $c$ represents the context, and $x_0^w$ is the preferred (winning) generated output over $x_0^l$ (losing). The method implicitly optimizes a reward function by ensuring the model prefers $x_0^w$, while remaining close to the original reference distribution, $p_{\text{ref}}(x_0 | c)$. The training objective minimizes the negative log-likelihood of the preference data using a ratio-based formulation:
\begin{equation}
    \label{eq_dpo}
    \begin{split}
        \mathcal{L}_\mathrm{\text{DPO}}(\theta)\!\!=\!\!-\!\mathbb{E}_{x_0^w\!, x_0^l, c} \!\left[\!\log\!\sigma\!\left(\!\!\beta\!\! 
        \left(\!\!\log\!{\frac{p_\theta(x_0^w\!|c)}{p_{\text{ref}}(x_0^w\!|c)}}\!-\!\log\!{\frac{p_\theta(x_0^l\!|c)}{p_{\text{ref}}(x_0^l\!|c)}}\!\!\right)\!\!\!\right)\!\!\right]\!\!,
    \end{split} 
\end{equation}
where the term inside the sigmoid $\sigma$ represents the difference in implicit rewards assigned to the winning and losing samples. The coefficient $\beta$ controls how strictly the optimization is anchored to the reference model $p_{\text{ref}}$. The intuition behind $\mathcal{L}_\mathrm{\text{DPO}}$ becomes clearer when analyzing how the parameters are updated during training. If we define the implicit reward for a sample as $\hat{r}_{\theta}(x_0,c)=\beta\cdot\log{\frac{p_\theta(x_0|c)}{p_{\text{ref}}(x_0|c)}}$, the derivative of the loss function with respect to the parameters $\theta$ of the model is the following:
\begin{equation}
    \label{eq_grad_dpo}
    \begin{split}
        \frac{\partial \mathcal{L}_\mathrm{\text{DPO}}(\theta)}{\partial\theta}\!&=\! -\beta \mathbb{E}_{x_0^w, x_0^l, c}\Bigg[ \sigma\left( \hat{r}_\theta(x_0^l, c) - \hat{r}_\theta(x_0^w, c)\right)\cdot\\ &\left( \frac{\partial\log{p_\theta(x_0^w|c)}}{\partial\theta} - \frac{\partial\log{p_\theta(x_0^l|c)}}{\partial\theta} \right)\!\Bigg].
    \end{split}
\end{equation}
Eq.~\eqref{eq_grad_dpo} has two key properties. First, the gradient direction acts to increase the probability of the preferred output $x_0^w$, while suppressing the disfavored $x_0^l$. Second, the magnitude of the update is dynamically scaled by the sigmoid term $\sigma\left( \hat{r}_\theta(x_0^l, c) - \hat{r}_\theta(x_0^w, c)\right)$, acting like a weighting factor. It approaches 1 when the model incorrectly assigns a higher reward to the losing sample, making the update strong, and vanishes towards 0 when the model correctly ranks the two examples.

\noindent
\textbf{Diffusion-DPO}. To apply DPO to the continuous domain of diffusion, Wallace~\etal~\cite{Wallace-arxiv-2023} replaced the log-probability terms with MSE objectives. The resulting loss penalizes the model if the denoising error on the preferred sample $x_t^w$ is not significantly lower than the error on the rejected sample $x_t^l$, relative to the reference model. The objective is defined as:
\begin{equation}
    \label{eq_diffusion_dpo}
    \begin{split}
        \!\mathcal{L}_{\mathrm{\mbox{\scriptsize{Diff-DPO}}}}&(\theta) \!=\! - \mathbb{E}_{x_t^w, x_t^l, c}\!\Big[\!\log\sigma\Big(\!\!-\!\beta \cdot T 
        \Big(\\ &
    \lVert\epsilon^w\!-\!\epsilon_\theta^w(x_t^w\!, t, c)\rVert_2^2\!-\! 
    \lVert\epsilon^w\!-\!\epsilon_{\text{ref}}^w(x_t^w, t, c)\rVert_2^2 -\\ &
    \lVert\epsilon^l\!-\!\epsilon_\theta^l(x_t^l, t, c)\rVert_2^2\!+\! 
    \lVert\epsilon^l\!-\!\epsilon_{\text{ref}}^l(x_t^l, t, c)\rVert_2^2
    \!\Big)\!\Big)\!\Big]\!,
    \end{split}
\end{equation}
where $x_t^w$ and $x_t^l$ represent the diffusion samples at timestep $t$, obtained by adding noise to the original pair $(x_0^w, x_0^l)$.

\vspace{-0.2cm}
\subsection{Consistency-DPO} 
\vspace{-0.1cm}

We extend the Diffusion-DPO formulation of Wallace \etal~\cite{Wallace-arxiv-2023} to consistency models. Instead of using the denoising error, we leverage the consistency distillation loss in Eq.~\eqref{eq_consistency_distillation} as a proxy for sample quality. The core principle of our approach is to optimize the parameters of the model such that the consistency mapping error is minimized for preferred examples $x_{t_{n}}^w$ compared with dispreferred examples $x_{t_{n}}^l$. Thus, we define the training objective as:
\begin{equation}
    \label{eq_consistency_dpo}
    \begin{split}
        \mathcal{L}_\mathrm{\text{Con-DPO}}&(\phi)\!=\! -\mathbb{E}_{x_{t_{n+1}}^w, x_{t_{n+1}}^l, c} \Big[\log \sigma\Big(\!-\beta\Big(\\ & d^w(x_{t_{n+1}}^w,\hat{x}_{t_n}^{w,\theta}, \phi) - d^l(x_{t_{n+1}}^l,\hat{x}_{t_n}^{l,\theta}, \phi)\!\Big)\!\Big)\!\Big], 
    \end{split}
\end{equation}
where, for $* \in \{w, l\}$, the distance $d^*$ is defined as:
\begin{equation}
    \label{eq_d_star}
    \begin{split}
       d^* = &\,d(f_\phi(x_{t_{n+1}}^*, t_{n+1}, c), f_{\text{ref}}(\hat{x}_{t_{n}}^{*,\theta}, t_{n}, c)) - \\ & \,d(f_{\text{ref}}(x_{t_{n+1}}^*, t_{n+1}, c), f_{\text{ref}}(\hat{x}_{t_{n}}^{*,\theta}, t_{n}, c)). 
    \end{split}
\end{equation}
The input $x_{t_{n+1}}^*$ is derived from $x_0^*$ (where $* \in \{w, l\}$) via the standard forward process. The target estimate $\hat{x}_{t_n}^{*,\theta}$ is computed by advancing $x_{t_{n+1}}^*$ one step along the PF-ODE using a pre-trained diffusion model $\theta$. The optimization is based on the distance metric $d$ and sigmoid $\sigma$, with indices $n \sim \mathcal{U}(1, N)$ spanning the discretization of the time interval.
The active parameters of the consistency model are represented by $\phi$. We diverge from standard practice by excluding the parameter running average ($\phi^-$), as the frozen reference consistency model $f_{\text{ref}}$ already has the self-consistency property.

\begin{algorithm*}[!t]
\caption{Curriculum-DPO++ (for consistency models)}
\label{alg:method}
\KwIn{$\{(x_{0, i}, c)\}_{i=1}^M$ - the training samples, $r_\varphi(x_0, c)$ - the reward model which can be conditioned on $c$, $B$ - the number of batches for splitting the set of pairs, $\theta$ - the parameters of a pre-trained diffusion model, $\alpha_t, \sigma_t$ - the parameters of the noise schedule, $T$ - the last diffusion timestep, $N$ - the discretization length of the interval $[0, T]$, $\beta$ - DPO hyperparameter to control the divergence from the initial pre-trained state, $\sigma$ - the sigmoid function, $d^*$ - as defined in Eq.~\eqref{eq_d_star}, $\eta$ - the learning rate, $\{H_k\}_{k=1}^B$ - the number of training iterations after including the $k$-th batch, $\{G_k\}_{k=1}^B$\ - the trainable layers after including the $k$-th batch, $\phi$ - the pre-trained weights of the reference model, 
$\mathbf{r}_{\text{start}}$ - the rank of the LoRA adaptation matrices at the start of the training, $\mathbf{r}_{\text{end}}$ - the rank of the LoRA adaptation matrices at the end of the training, $\delta$ - the rank update rate.
}
\KwOut{
$\phi$ - the trained weights of the consistency model.}
$\hat{X} \leftarrow \{(x_{0,i}, c)|r_\varphi(x_{0,i}, c) \leq r_\varphi(x_{0,i-1},  c), i=\{2,3, \dots ,M\}\};$ \textcolor{commentGreen}{$\lhd$ sort the samples in descending order of the rewards}\\

$S \leftarrow \left\{(x_{0,i}, x_{0, j}, c)| i,j \in \{1, \dots, M\}; i<j; x_{0,i}, x_{0, j} \in \hat{X},  r_\varphi(x_{0,i}, c) > r_\varphi(x_{0,j}, c)  \right\}$; \textcolor{commentGreen}{$\lhd$ create pairs of examples using the order from $\hat{X}$}\\
$\mathrm{L}_k \leftarrow \left\{\frac{(M-1) \cdot(B-k)}{B}\right\}_{k=1}^B$; \textcolor{commentGreen}{$\lhd$ the minimum preference limits of the batches}\\
$\mathrm{R}_k \leftarrow \left\{\frac{(M-1)\cdot(B-(k-1))}{B}\right\}_{k=1}^B$;  \textcolor{commentGreen}{$\lhd$ the maximum preference limits of the batches}\\
$S_k\leftarrow \left\{(x_{0}^w, x_{0}^l, c) | (x_{0}^w, x_{0}^l) = (x_{0,i}, x_{0, j}); \mathrm{L}_k   < j-i \leq \mathrm{R}_k  ; (x_{0,i}, x_{0, j}, c) \in S \right\}_{k = 1}^B$;  \textcolor{commentGreen}{$\lhd$ the batches of increasingly difficult pairs}\\

$\mathbf{r} \leftarrow \left\{\mathbf{r}_k|k \in \{1, \dots , B\}, \mathbf{r}_1 = \mathbf{r}_{\text{start}}, \mathbf{r}_{k+1} = \min\left(\mathbf{r}_{\text{end}}, \mathbf{r}_{k} \cdot \delta\right)\right\}$; \textcolor{commentGreen}{$\lhd$ the ranks of the low-rank matrices in LoRA}\\

$\text{LoRA}_k = \left\{\left(A^{\mathbf{r}_k}_i, C^{\mathbf{r}_k}_i\right) | i \in G_k,A^{\mathbf{r}_k}_i \sim \mathcal{N}(0,\sigma), C^{\mathbf{r}_k}_i = \mathbf{0}_{m\times \mathbf{r}_k} \right\}_{k=1}^B$; \textcolor{commentGreen}{$\lhd$ initialization of LoRA matrices} \\

$P \leftarrow \emptyset$;  \textcolor{commentGreen}{$\lhd$ current training set} \\

$G \leftarrow \emptyset$; \textcolor{commentGreen}{$\lhd$ current trainable layers}\\

\ForEach{$k \in \{1, \dots, B\}$}
{
    $P \leftarrow P \cup S_k$; \textcolor{commentGreen}{$\lhd$ include a new batch in the training}\\
    $G^{\text{new}} \leftarrow \text{LoRA}_k$; \textcolor{commentGreen}{$\lhd$ include a new set of trainable layers}\\
    \ForEach{ $\left(A^{\mathbf{r}_k}_i, C^{\mathbf{r}_k}_i\right) \in G^{\textnormal{new}}$}
    {
        \If{$i < |G|$}{ \textcolor{commentGreen}{//  transfer the previously learned weights to the new LoRA matrices} \\
        $A^{\mathbf{r}_{k-1}}_i, C^{\mathbf{r}_{k-1}}_i \leftarrow G_i $;\\
        $A^{\mathbf{r_{k}}}_i\left[:\mathbf{r}_{k-1}, :\right] \leftarrow A^{\mathbf{r}_{k-1}}_i$; \\

        $C^{\mathbf{r_{k}}}_i\left[:,:\mathbf{r}_{k-1}\right] \leftarrow C^{\mathbf{r}_{k-1}}_i$; \\
        
        }

    }
    $G \leftarrow G^{\text{new}}$; \\
    \ForEach{$i \in \{1, \dots, H_k\}$}
    {
    $(x_0^w, x_0^l, c) \sim \mathcal{U}(P)$; 
    $n \sim \mathcal{U}[1,N-1]$; $\epsilon \sim \mathcal{N}(0, \mathbf{I})$;\\
    $x_{t_{n+1}}^w \leftarrow \alpha_{t_{n+1}} x_0^w + \sigma_{t_{n+1}} \epsilon$;  \textcolor{commentGreen}{$\lhd$ forward process} \\
    $x_{t_{n+1}}^l \leftarrow \alpha_{t_{n+1}} x_0^l + \sigma_{t_{n+1}} \epsilon$; \textcolor{commentGreen}{$\lhd$ forward process}\\
    $\hat{x}_{t_{n}}^{w,\theta} \leftarrow \mathrm{ODESolver}(x_{t_{n+1}}^w, \theta, \alpha_{t_{n+1}}, \sigma_{t_{n+1}})$; \textcolor{commentGreen}{$\lhd$ one denoising step} \\
    $\hat{x}_{t_{n}}^{l,\theta} \leftarrow \mathrm{ODESolver}(x_{t_{n+1}}^l, \theta, \alpha_{t_{n+1}}, \sigma_{t_{n+1}})$; \textcolor{commentGreen}{$\lhd$ one denoising step}\\
    $\mathcal{L}_\mathrm{\text{Con-DPO}}(\phi, G) \leftarrow - \log \sigma\Big(-\beta( d^w(x_{t_{n+1}}^w,\hat{x}_{t_n}^{w,\theta}, \phi, G) - d^l(x_{t_{n+1}}^l,\hat{x}_{t_n}^{l,\theta}, \phi,G))\Big)$; \textcolor{commentGreen}{$\lhd$ DPO loss, as in Eq.~\eqref{eq_consistency_dpo}}\\
    $G \leftarrow G - \eta \frac{\partial \mathcal{L}_\mathrm{\text{Con-DPO}}}{\partial G}$; \textcolor{commentGreen}{$\lhd$ update the weights}
    }
}
$\phi \leftarrow \phi + G$; \textcolor{commentGreen}{$\lhd$ store the LoRA weights into the original model}\\
\textbf{return} $\phi$
\end{algorithm*}

To validate that Consistency-DPO shares the optimization dynamics of the original DPO (Eq.~\eqref{eq_dpo}) and Diffusion-DPO (Eq.~\eqref{eq_diffusion_dpo}), we further examine the gradient of the loss with respect to the parameters $\phi$:
\begin{equation}
    \label{eq_grad_consistency_dpo}
    \begin{split}
        \frac{\partial \mathcal{L}_\mathrm{\text{Con-DPO}}}{\partial \phi}\!=\!  \beta\mathbb{E}_{x_{t_{n+1}}^w\!, x_{t_{n+1}}^l\!, c}\!\Bigg[\!\sigma\!\left(\beta\!\left(d^w\!-\!d^l\right)\!\right)\!\!\left(\!\frac{\partial d^w}{ \partial \phi}\!-\!\frac{\partial d^l}{\partial \phi}\!\right)\!\!\!\Bigg]\!.
    \end{split}
\end{equation}
Eq.~\eqref{eq_grad_consistency_dpo} reduces the consistency metric $d^w$ for the favored samples, while increasing $d^l$ for the less preferred examples. The strength of this correction is dynamically weighted by $\sigma(d^w-d^l)$, which approaches 1 when $f_\phi$ exhibits a weaker preference for $x^w$ compared with $f_{\text{ref}}$, ensuring that the model learns primarily from misranked pairs.

\vspace{-0.2cm}
\subsection{Curriculum-DPO}
\vspace{-0.1cm}
In our preliminary work \cite{Croitoru-CVPR-2025}, we challenge the assumption that all preference pairs are equally informative at all stages of training. We observe that preference is hierarchical, \ie~some choices are obvious, for example broken versus coherent images, while others are subtle. To exploit this, we implement a coarse-to-fine learning process. The model is initially trained on pairs with high-level semantic discrepancies, allowing it to rapidly grasp fundamental preference directions. As training advances, the curriculum shifts toward pairs requiring fine-grained discrimination. Formally, we propose a curriculum guided by the reward model $r_\varphi(x_0, c)$, which serves as a proxy for establishing ground-truth rankings. By leveraging the score differential between the winner and loser in a pair $(x_0^w, x_0^l)$, we can estimate the distinctiveness of the preference signal. Wide gaps in scoring suggest easy pairs, where the superior sample is visually distinct due to coarse-level factors. Narrow gaps indicate hard pairs, where the preference relies on subtle details. We also use a lower-bound cutoff on this difference. This ensures the model ignores pairs where the preference decision is ambiguous, thereby mitigating the risk of fitting to the reward model's biases, rather than true signal.

In Algorithm~\ref{alg:method}, we outline the Curriculum-DPO++ procedure. By ignoring steps 6-7 and 12-19, we obtain the original Curriculum-DPO. Initially, $M$ samples are sorted by a reward score $r_\varphi$ (step 1). Next, we partition the set of valid preference pairs into $B$ difficulty levels (subsets $S_k$), determined by the rank difference between paired samples bounded by $[\mathrm{L}_k, \mathrm{R}_k]$. Training proceeds progressively (steps 11-27). At each stage $k$, the corresponding subset $S_k$ is merged into the active training set $P$, followed by $H_k$ optimization steps. The process starts with $S_1$, representing the pairs with the most obvious preference signals. The last part of the algorithm (steps 21-27) comprises steps specific to Consistency-DPO. The corresponding derivation for Diffusion-DPO is available in Appendix \ref{supp_alg}.

\vspace{-0.2cm}
\subsection{Curriculum-DPO++}
\vspace{-0.1cm}

When preference patterns are evident, a shallow model should be sufficient to distinguish between winning and losing samples. As the complexity of training samples increases during training, the learning capacity of the model must expand to cope with the harder preference learning task. However, using a model with fixed learning capacity is suboptimal, since a high-capacity model can overfit on easy examples, while a low-capacity model might underfit on hard examples. To address the misalignment between model capacity and task complexity, we complement our data-level curriculum with a new model-level curriculum that progressively scales the complexity of the network along with the difficulty of the preference pairs. Since the fine-tuning is based on LoRA~\cite{Hu-ICLR-2022}, we propose to grow the learning capacity of the model during training via dynamically increasing (i) the rank of the low-rank matrices and (ii) the number of trainable layers.

In Algorithm~\ref{alg:method}, we detail the Curriculum-DPO++ training process for consistency models, which includes the process of scaling the learning capacity of the model when introducing more complex preference pairs. We establish two schedules synchronized with the introduction of each data-level curriculum batch $S_k$, $\forall k=\{1, \dots, B\}$. First, we define the layer schedule $\left\{G_k\right\}_{k=1}^B$, where $G_k$ denotes the set of active trainable layers for the $k$-th data batch, such that $G_k$ contains more and more layers as $k$ grows. In practice, at each iteration, we add one upsampling and one downsampling block to the new set of trainable layers. Second, we govern the model capacity via a rank schedule for the LoRA matrices. The standard LoRA approximates the weight updates via two fixed-size low-rank matrices, $C \in \mathbb{R}^{m\times r}$ and $A \in \mathbb{R}^{r \times h}$, where $h$ is the size of the input features, $m$ is the size of the output features, and $r \ll \min\left(m, h\right)$ determines the rank of the two matrices. In our algorithm, the rank $\mathbf{r}_k$ is not fixed during training, as it changes according to the hyperparameters $\mathbf{r}_{\text{start}}$, $\mathbf{r}_{\text{end}}$, and the growth rate $\delta$, as formalized in step 6 of Algorithm~\ref{alg:method}. However, we emphasize that the chosen values still respect the low-rank property $\mathbf{r}_k \ll \min\left(m, h\right)$. In step 7, we initialize all the matrices that are to be used throughout the training, but in practice, we can initialize these matrices exactly when we need them. Before entering the training loop on curriculum batch $S_k$, we establish the trainable layers in step 12 of the algorithm. If $k>1$, we transfer the weights from the previously learned LoRA matrices to the new ones in steps 13-18. In step 28, the final LoRA weights are integrated into the pre-trained consistency model $\phi$, leading to a preference-tuned version of the respective model.

\vspace{-0.2cm}
\subsection{Reward-Model-Free Curriculum}
\vspace{-0.1cm}

As an orthogonal contribution, we introduce a curriculum-based training methodology that eliminates the need for an explicit reward model. Instead of assigning external scores to each image, we induce implicit preference signals by perturbing the prompt embeddings used for image generation. Specifically, with the text alignment task in mind, we apply a masking operation on the prompt embedding, \ie~zero masking. Higher masking ratios disrupt the semantic information provided to the generator, resulting in images that exhibit weaker adherence to the input prompt, and sometimes, reduced synthesis quality. With these degraded samples, we can naturally form winning-losing pairs for DPO fine-tuning, where unmasked prompt embeddings serve to generate preferred outcomes.

To incorporate a curriculum strategy, we vary the masking ratio from $0.9$ to $0.1$. At the start of training, winning samples correspond to images generated from unmasked prompts, while losing samples are drawn from prompts with a masking ratio of $0.9$. Over time, we progressively introduce losing samples with smaller masking ratios, enabling the model to learn increasingly subtle distinctions in text-image alignment. It is important to note that we always use unmasked examples
as winning samples. Therefore, the easy-to-hard curriculum is created by decreasing the masking ratio from $0.9$ to $0.1$ for the losing examples. 

\vspace{-0.2cm}
\section{Experiments}
\label{sec: experiments}

\begin{table*}[t]

  \centering
  \setlength\tabcolsep{0.46em}
  \small{
  \begin{tabular}{llcccccc}
  \toprule
   \multirow{3.5}{*}{Dataset}   & \multirow{3.5}{*}{Fine-Tuning Strategy}     & \multicolumn{3}{c}{LCM}   & \multicolumn{3}{c}{SD} \\ 
    \cmidrule(l{2pt}r{2pt}){3-5}
    \cmidrule(l{2pt}r{2pt}){6-8}
         &      & Text  & \multirow{2}{*}{Aesthetics} & Human   & Text & \multirow{2}{*}{Aesthetics} & Human  \\  
         &      &  Alignment &  & Preference  &  Alignment &  &  Preference \\
    \midrule
    \multirow{5}{*}{$D_1$ \cite{Black-ICLR-2024}} & - & 0.7243$_{\pm0.0048}$  & 6.0490$_{\pm0.0162}$ & 0.2912$_{\pm0.0021}$  & 0.6804$_{\pm0.0052}$ & 5.5152$_{\pm0.0166}$ & 0.2784$_{\pm0.0015}$ \\
    &  DDPO \cite{Black-ICLR-2024} & 0.7490$_{\pm0.0036}$ & 6.3730$_{\pm0.0130}$ & 0.2952$_{\pm0.0011}$ & 0.7629$_{\pm0.0040}$ &
  5.8183$_{\pm0.0129}$ &
  0.2854$_{\pm 0.0012}$ \\

    &  DPO \cite{Wallace-arxiv-2023} & 0.7502$_{\pm 0.0045}$ & 6.4741$_{\pm0.0095}$ & 0.2990$_{\pm 0.0010}$ &  0.7614$_{\pm0.0049}$ & 5.7146$_{\pm0.0162}$ & 0.2827$_{\pm0.0008}$\\

    & Curriculum-DPO~\cite{Croitoru-CVPR-2025} & 0.7548$_{\pm0.0041}$ & 6.6417$_{\pm0.0083}$& 0.3237$_{\pm0.0012}$ & 0.7703$_{\pm0.0036}$ & 5.8232$_{\pm0.0197}$ & 0.2856$_{\pm0.0015}$ \\
    
    & Curriculum-DPO++ & \textbf{0.7603}$_{\pm0.0013}$ & \textbf{6.7065}$_{\pm0.0080}$&
    \textbf{0.3241}$_{\pm0.0012}$ &
    \textbf{0.7744}$_{\pm0.0037}$ & \textbf{5.8342}$_{\pm0.0125}$ &
     \textbf{0.3325}$_{\pm0.0008}$ \\

  \midrule
  \multirow{5}{*}{$D_2$ \cite{Saharia-NeurIPS-2022}} & -  & 0.5602$_{\pm0.0032}$ &  5.8038$_{\pm0.0139}$ & 0.2610$_{\pm0.0016}$  & 0.5997$_{\pm0.0067}$ &  5.4292$_{\pm0.0181}$ &
  0.2646$_{\pm0.0011}$ \\
  & DDPO \cite{Black-ICLR-2024} & 0.5721$_{\pm0.0043}$  & 6.0121$_{\pm0.0127}$  & 0.2803$_{\pm0.0019}$ &
  0.6024$_{\pm0.0055}$ &
  5.6748$_{\pm0.0162}$ &
  0.2673$_{\pm0.0025}$ \\
    
  &  DPO \cite{Wallace-arxiv-2023} & 0.5720$_{\pm0.0040}$ & 6.0430$_{\pm0.0113}$ & 0.2814$_{\pm0.0023}$  & 0.6075$_{\pm0.0057}$ &
  5.6205$_{\pm0.0152}$ &
  0.2672$_{\pm0.0013}$ \\

    & Curriculum-DPO~\cite{Croitoru-CVPR-2025} &  0.5812$_{\pm0.0038}$ & 6.1829$_{\pm0.0128}$ & 0.2851$_{\pm0.0017}$ & 0.6234$_{\pm0.0050}$ & 5.7060$_{\pm0.0177}$ & 0.2681$_{\pm0.0019}$ \\

     & Curriculum-DPO++ & \textbf{0.5858}$_{\pm0.0025}$ &
    \textbf{6.3626}$_{\pm0.0622}$  &
    \textbf{0.2903}$_{\pm0.0018}$ &
     \textbf{0.6247}$_{\pm0.0014}$ &
     \textbf{5.8890}$_{\pm0.0083}$ &
     \textbf{0.2955}$_{\pm0.0014}$ \\
\midrule
\multirow{5}{*}{$D_3$~\cite{Kirstain-NeurIPS-2023}}&-  & 0.5091$_{\pm0.0064}$
& 6.1522$_{\pm0.1019}$
& 0.2798$_{\pm0.0016}$ & 0.5270$_{\pm0.0022}$ &
   5.6713$_{\pm0.0034}$ &
   0.2687$_{\pm0.0013}$  \\
   & DDPO~\cite{Black-ICLR-2024} & 0.5199$_{\pm0.0046}$ &
   5.9874$_{\pm0.0218}$&
  0.2831$_{\pm0.0011}$ & 0.5344$_{\pm0.0036}$ &
   5.7066$_{\pm0.0279}$ &
   0.2708$_{\pm0.0009}$
   \\
   & DPO~\cite{Wallace-arxiv-2023} 
   & 0.5158$_{\pm0.0053}$
   & 6.2790$_{\pm0.1282}$
   & 0.2850$_{\pm0.0049}$  
   & 0.5369$_{\pm0.0084}$ & 
   5.7644$_{\pm0.0046}$ & 
   0.2736$_{\pm0.0016}$\\
     & Curriculum-DPO~\cite{Croitoru-CVPR-2025} 
     & 0.5230$_{\pm0.0011}$ 
     & 6.4337$_{\pm0.0251}$ 
     & 0.3011$_{\pm0.0034}$ &
     0.5230$_{\pm0.0011}$ 
     & 6.4337$_{\pm0.0251}$ 
     & 0.3011$_{\pm0.0034}$\\
      & Curriculum-DPO++ & \textbf{0.5236}$_{\pm0.0029}$ &
      \textbf{6.4572}$_{\pm0.0361}$ &
      \textbf{0.3050}$_{\pm0.0005}$  & \textbf{0.5236}$_{\pm0.0029}$ &
      \textbf{6.4572}$_{\pm0.0361}$ &
      \textbf{0.3050}$_{\pm0.0005}$ \\
  \bottomrule
  \end{tabular}
  }\vspace{-0.2cm}
    \caption{Text alignment, aesthetic and human preference scores on datasets $D_1$, $D_2$ and $D_3$, obtained by the baseline (pre-trained) LCM and SD models versus the four fine-tuning strategies: DDPO, DPO, Curriculum-DPO and Curriculum-DPO++. The results represent averages over three runs. The best scores on each dataset are highlighted in bold.}
  \label{tab_strategies}
  \vspace{-0.3cm}
\end{table*}

\begin{figure*}[t]
  \centering
  \includegraphics[width=0.9\linewidth]{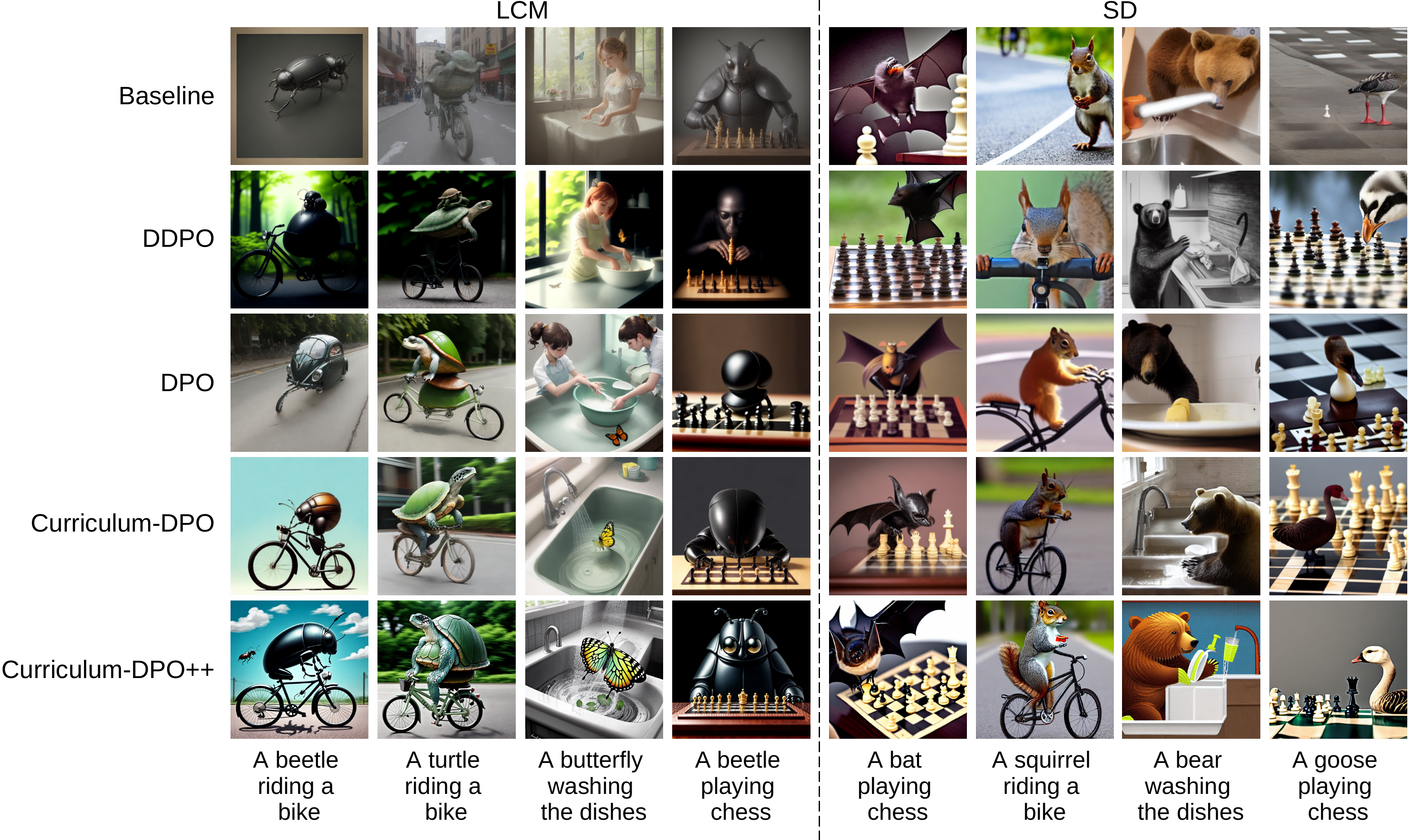}
  \vspace{-0.2cm}
   \caption{Qualitative results on dataset $D_1$, before and after fine-tuning for the text alignment task. The fine-tuning alternatives are DDPO, DPO, Curriculum-DPO, and Curriculum-DPO++. Best viewed in color.}
   \vspace{-0.3cm}
   \label{qualitative_text_align}
\end{figure*}

\vspace{-0.1cm}
\textbf{Datasets.} We conduct experiments across three datasets, denoted as $D_1$, $D_2$ and $D_3$. We adopt the data generation methodology of Black \etal~\cite{Black-ICLR-2024} to construct the first dataset ($D_1$). For text alignment and human preference, we utilize Subject-Verb-Object (SVO) prompts (\eg~``a dog washing the dishes``) derived from a combination of 45 animal classes and 3 activity types. For the aesthetics task, we use the template ``a photo of $<$an object$>$``, sampling from the same 45 animal classes. By generating 500 images per prompt, we obtain 67,500 training and 6,750 evaluation pairs for the alignment and preference tasks, while the aesthetics task comprises 22,500 training and 2,250 evaluation pairs. 

Dataset $D_2$ is based on DrawBench~\cite{Saharia-NeurIPS-2022}. By generating 500 images for each of its 200 prompts, we construct a set of 100,000 pairs for training. For evaluation, we generate 50 images per prompt, resulting in a total of 10,000 images.

Dataset $D_3$ consists of 150,000 pairs from Pick-a-Pic~\cite{Kirstain-NeurIPS-2023}, a dataset where images are already provided. For Pick-a-Pic, we evaluate on 500 official test prompts.



\noindent
\textbf{Generative models.} 
Our experiments leverage two pre-trained generative models, namely the Stable Diffusion (SD) model~\cite{rombach-CVPR-2022} and the Latent Consistency Model (LCM)~\cite{Luo-arXiv-2023}. The LCM checkpoint is derived via consistency distillation from SD v1.5. For SD, we generate images of $256\times256$ pixels using 50 DDIM steps, whereas for LCM, we generate image at a resolution of $768\times768$ pixels using 8 steps of Multistep Latent Consistency Sampling~\cite{Luo-arXiv-2023}.


\noindent
\textbf{Reward models}. 
Following the reward modeling protocol of Black \etal~\cite{Black-ICLR-2024}, we employ three metrics to guide and evaluate our models. To measure text alignment, we compute the cosine similarity between Sentence-BERT~\cite{reimers-EMNLP-2019} embeddings of the original prompt and a caption generated by LLaVA~\cite{liu-NeurIPS-2024}. For visual appeal, we utilize the LAION Aesthetics Predictor~\cite{Schuhmann-laion-2022}, a linear model based on CLIP~\cite{radford-ICML-2021} representations. Finally, human preference is estimated using HPSv2~\cite{Wu-arXiv-2023}, a CLIP model fine-tuned on human-ranked image pairs. Crucially, DPO, DDPO, Curriculum-DPO and Curriculum-DPO++ utilize  identical reward models to ensure a fair and consistent comparison.


\noindent
\textbf{Competing methods}. We compare Curriculum-DPO++ with the original Curriculum-DPO~\cite{Croitoru-CVPR-2025}, as well as established fine-tuning techniques, including Direct Preference Optimization (DPO)~\cite{Rafailov-NeurIPS-2023, Wallace-arxiv-2023} and Denoising Diffusion Policy Optimization (DDPO)~\cite{Black-ICLR-2024}. Additionally, we include the pre-trained base models (LCM and SD) as reference points.

\noindent
\textbf{Training setup}. We execute all experiments on a single H100 GPU, training for 10,000 iterations with a batch size of 16 and two steps of gradient accumulation. Computationally, LCM experiments require approximately 64GB of VRAM and 48 hours to complete, whereas SD experiments utilize 36GB of VRAM and finish in approximately 24 hours. 

We optimize using AdamW with a fixed learning rate of $5 \cdot 10^{-5}$. LoRA is applied in all cases, configured with rank and scale values of $r\!=\!64$ and $\alpha\!=\!64$ for LCM, and $r\!=\!8$ and $\alpha\!=\!8$ for SD, respectively. For our model-level curriculum, we choose the hyperparameters for LCM as follows: $\mathbf{r}_{\text{start}}\!=\!4$, $\mathbf{r}_{\text{end}}\!=\!64$ and $\delta\!=\!2$. For SD, we use the following values: $\mathbf{r}_{\text{start}}\!=\!6$, $\mathbf{r}_{\text{end}}\!=\!16$ and $\delta\!=\!2$. The layer schedule $\{G_k\}_{k=1}^B$ is implemented via a progressive growth strategy. Training starts with the middle block and the innermost upsampling and downsampling layers. At each subsequent stage $k$, the set of trainable layers expands by incorporating an additional symmetrical pair of downsampling and upsampling blocks. For the reward-model-free experiments, we begin with losing samples having a $0.9$ masking ratio, and decrease it by $0.1$ every 1,000 steps.

We set $\beta\!=\!5000$ for Diffusion-DPO~\cite{Wallace-arxiv-2023}, and $\beta\!=\!200$ for Consistency-DPO. The number of curriculum batches is set to $B\!=\!5$ for both models. 
We define the iteration schedule as $H_i\!=\!K\!=\!400$ for the initial batches $i\!\in\!\{1, \dots, 4\}$. The final batch consumes the remaining budget ($H_5\!=\!10,\!000\!-\!4\!\cdot\!K$), ensuring that the total number of iterations matches the 10,000 iterations of the baselines, for a fair comparison. Regarding sampling, we select image pairs based on the magnitude of their preference score difference rather than their rank difference. This approach effectively mitigates issues arising from skewed preference score distributions.

\begin{figure}[t]
    \centering
    \begin{subfigure}{0.45\linewidth}
        \includegraphics[width=0.9\linewidth]{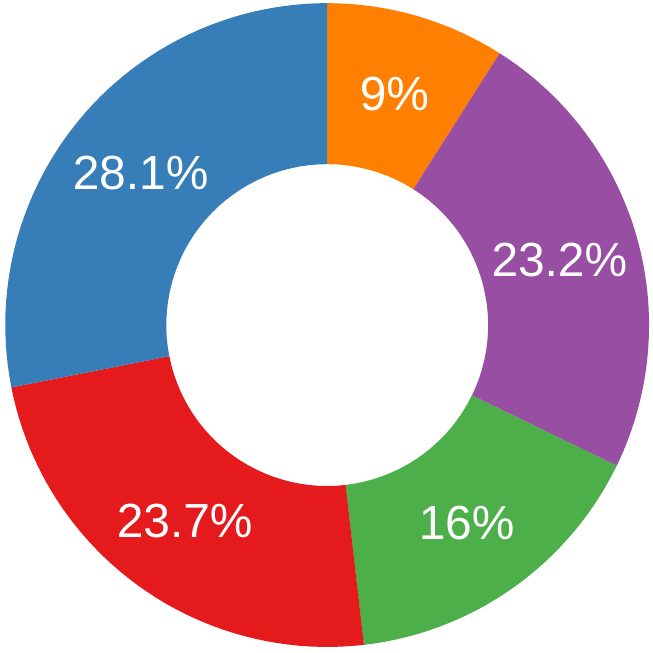}
        \vspace{-0.1cm}
        \caption{LAION win rates.}
        \vspace{-0.1cm}
    \end{subfigure}
    \begin{subfigure}{0.45\linewidth}
        \includegraphics[width=0.9\linewidth]{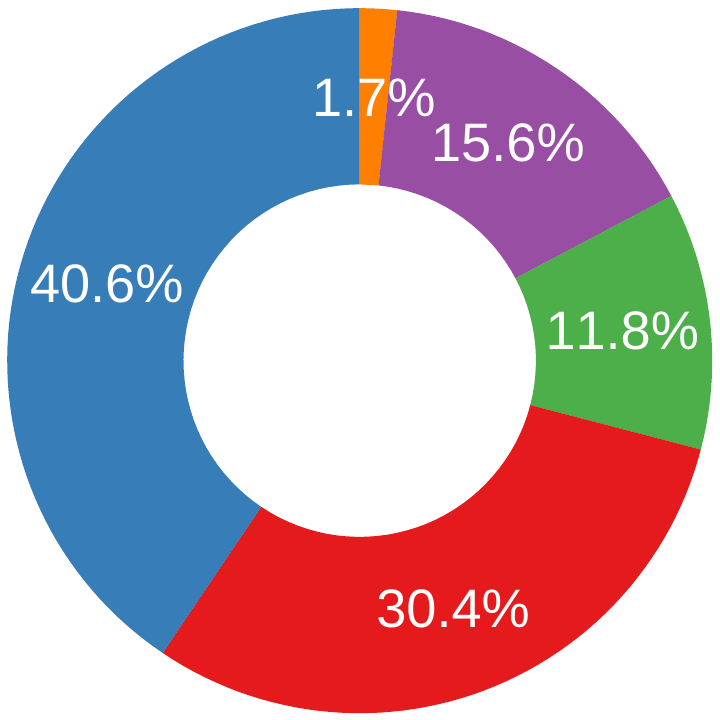}
        \vspace{-0.1cm}
        \caption{HPSv2 win rates.}
        \vspace{-0.1cm}
    \end{subfigure}
    \par\bigskip
    \begin{subfigure}{0.9\linewidth}
        \includegraphics[width=0.9\linewidth]{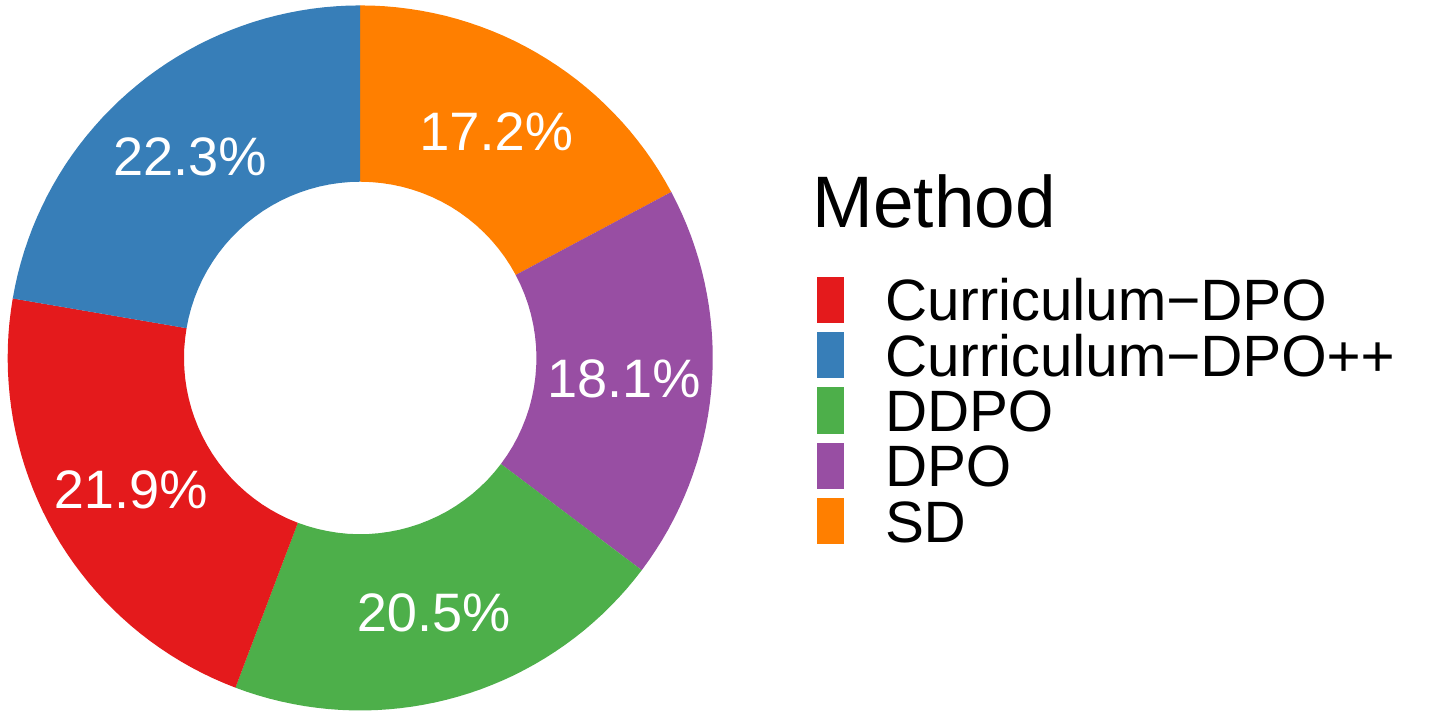}
        \vspace{-0.1cm}
        \caption{LLaVA win rates.}
        \vspace{-0.1cm}
    \end{subfigure}
    \vspace{-0.1cm}
    \caption{Preference win rates according to all the three reward models. The comparison is conducted between all the four fine-tuning strategies and the pre-trained SD baseline. For this comparison, we use images generated for the Pick-a-Pic ($D_3$) test set of prompts. Best viewed in color.}
    \label{fig_winrates_sd}
    \vspace{-0.4cm}
\end{figure}

\begin{figure}[t]
    \centering
    \begin{subfigure}{0.45\linewidth}
        \includegraphics[width=0.9\linewidth]{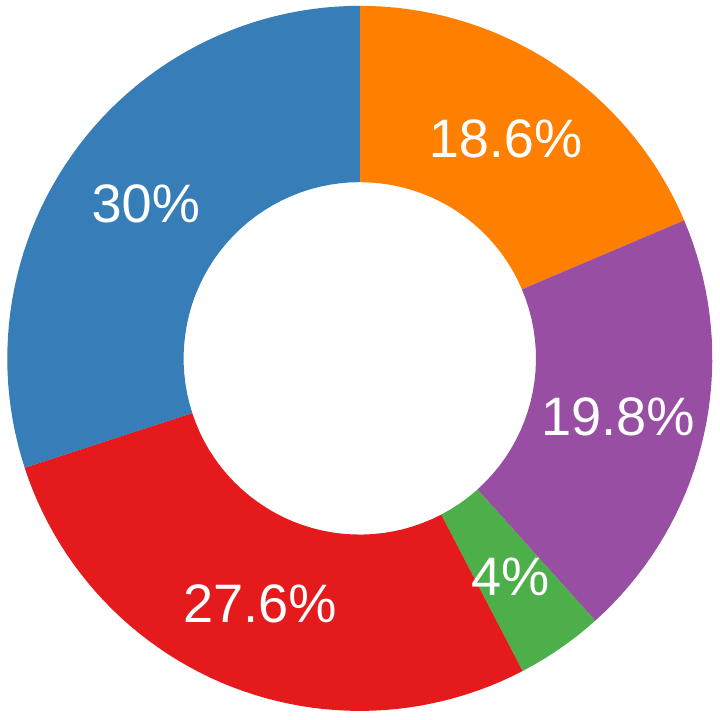}
        \vspace{-0.1cm}
        \caption{LAION win rates.}
        \vspace{-0.1cm}
    \end{subfigure}
    \begin{subfigure}{0.45\linewidth}
        \includegraphics[width=0.9\linewidth]{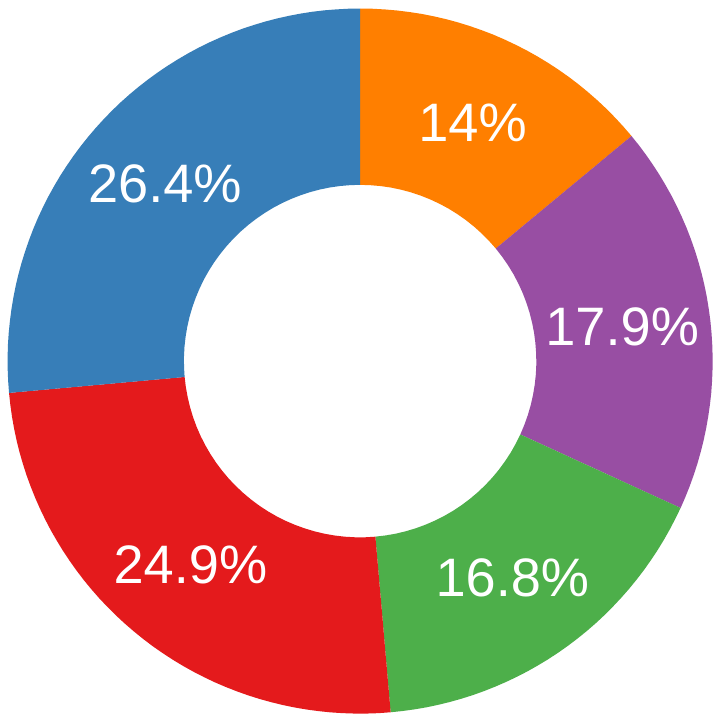}
        \vspace{-0.1cm}
        \caption{HPSv2 win rates.}
        \vspace{-0.1cm}
    \end{subfigure}
    \par\bigskip
    \begin{subfigure}{0.9\linewidth}
        \includegraphics[width=0.9\linewidth]{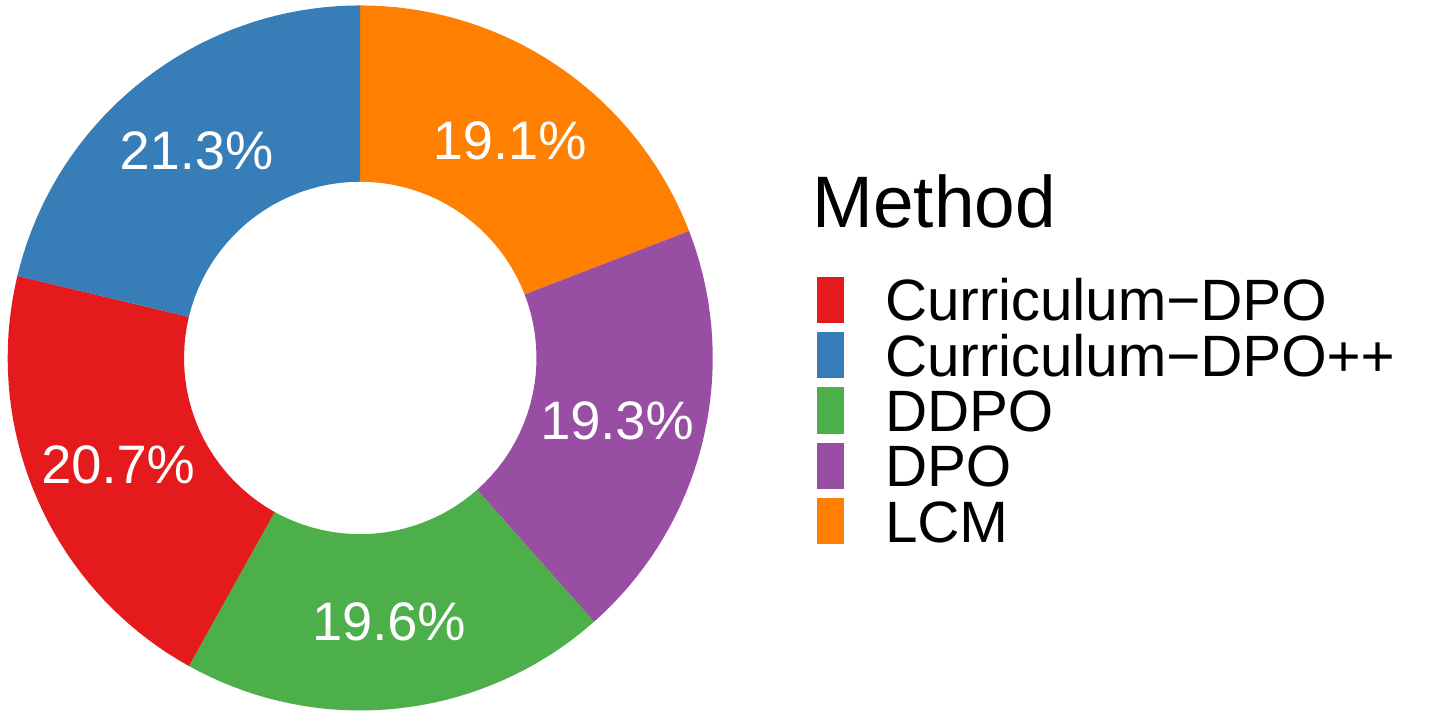}
        \vspace{-0.1cm}
        \caption{LLaVA win rates.}
        \vspace{-0.1cm}
    \end{subfigure}
    \vspace{-0.1cm}
    \caption{Preference win rates according to all the three reward models. The comparison is conducted between all the four fine-tuning strategies and the pre-trained LCM baseline. For this comparison, we use images generated for the Pick-a-Pic ($D_3$) test set of prompts. Best viewed in color.}
    \label{fig_winrates}
    \vspace{-0.4cm}
\end{figure}

\begin{table*}[t]
  \centering
  \setlength\tabcolsep{0.46em}
  \small{
  \begin{tabular}{llcccc}
  \toprule
     \multirow{3.5}{*}{Model} & \multirow{3.5}{*}{Fine-Tuning Strategy}      & \multicolumn{2}{c}{Dataset $D_1$ \cite{Black-ICLR-2024}}   & \multicolumn{2}{c}{Dataset $D_2$ \cite{Saharia-NeurIPS-2022}} \\ 
    \cmidrule(l{2pt}r{2pt}){3-4}
    \cmidrule(l{2pt}r{2pt}){5-6}
         &      & Text  & Human   & Text & Human  \\  
         &      &  Alignment & Preference & Alignment  &  Preference   \\
    \midrule
    \multirow{3}{*}{LCM} & - & 0.7243$_{\pm0.0048}$  & 0.2912$_{\pm0.0021}$ &  0.5602$_{\pm0.0032}$ & 0.2610$_{\pm0.0016}$ \\
    & Curriculum-DPO & 0.7224$_{\pm0.0010}$ & 0.2936$_{\pm0.0003}$ & 0.5667$_{\pm0.0017}$ & 0.2775$_{\pm0.0007}$ \\
    & Curriculum-DPO++ & \textbf{0.7264}$_{\pm0.0029}$ & \textbf{0.3003}$_{\pm0.0014}$ & \textbf{0.5577}$_{\pm0.0014}$ & \textbf{0.2809}$_{\pm0.0010}$ \\

  \midrule
  \multirow{3}{*}{SD} & -  & 0.6804$_{\pm0.0052}$ & 0.2784$_{\pm0.0015}$ &  0.5997$_{\pm0.0067}$ &
  0.2646$_{\pm0.0011}$ \\
     & Curriculum-DPO & 0.7455$_{\pm0.0022}$ & 0.2894$_{\pm0.0018}$ & 0.6242$_{\pm0.0029}$ & \textbf{0.2708}$_{\pm0.0018}$ \\
     & Curriculum-DPO++ & \textbf{0.7472}$_{\pm0.0020}$ & \textbf{0.2947}$_{\pm0.0011}$ & \textbf{0.6272}$_{\pm0.0032}$ & 0.2695$_{\pm0.0021}$ \\

  \bottomrule
  \end{tabular}
  }\vspace{-0.2cm}
    \caption{Text alignment and human preference scores on datasets $D_1$ and $D_2$, obtained by the baseline (pre-trained) LCM and SD models versus the reward-model-free Curriculum-DPO and Curriculum-DPO++. The best scores are highlighted in bold.}
  \label{tab_model_free}
  \vspace{-0.4cm}
\end{table*}

\captionsetup[subfigure]{margin=2.8em}
\begin{figure*}[!t]
\begin{subfigure}[t]{.245\textwidth}
  \centering
  \includegraphics[width=.99\linewidth]{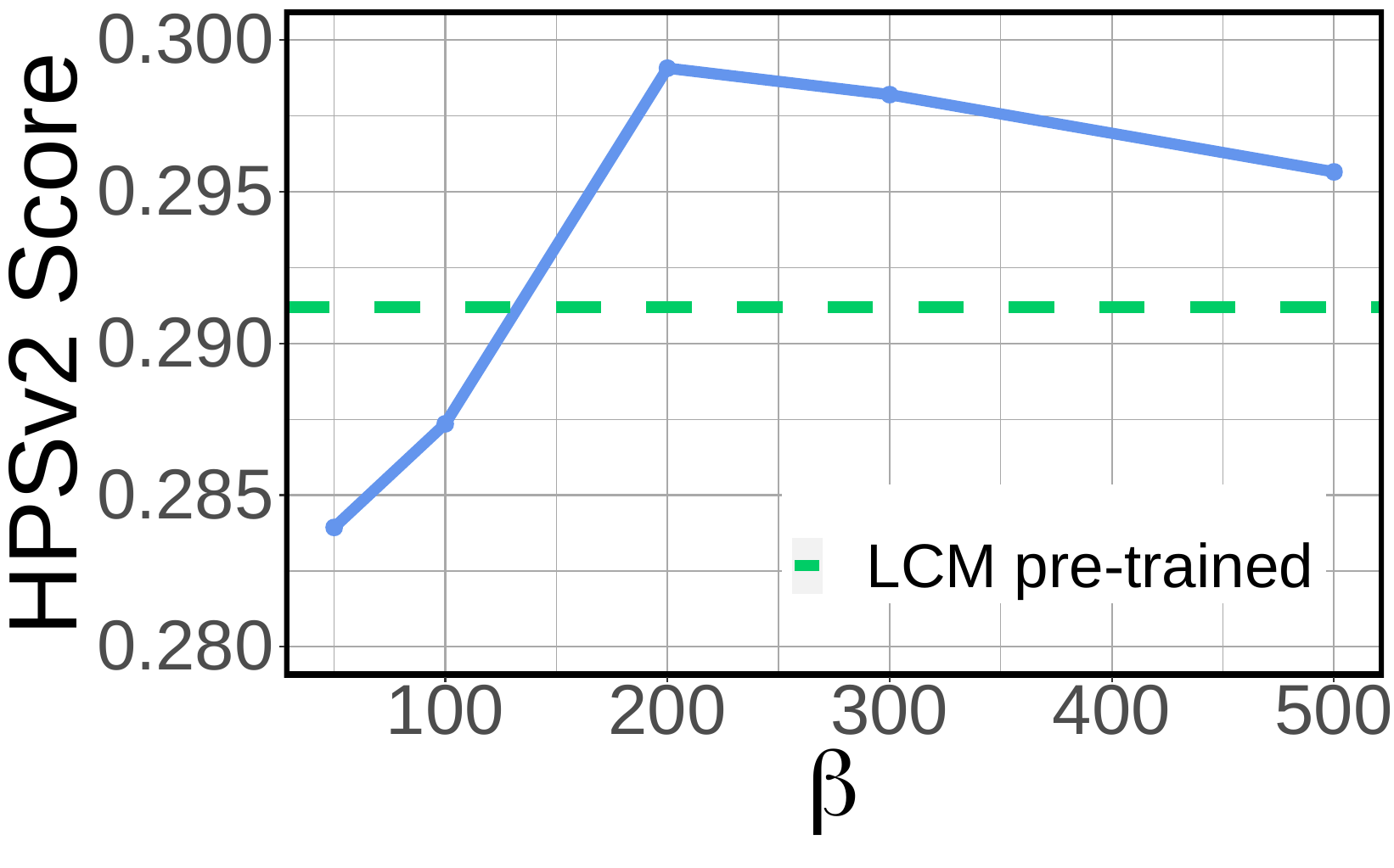}  
  \vspace{-0.6cm}
  \caption{Varying $\beta$ for Consistency-DPO.}
  \label{fig:sub-beta}
\end{subfigure}
\begin{subfigure}[t]{.245\textwidth}
  \centering
\includegraphics[width=.99\textwidth]{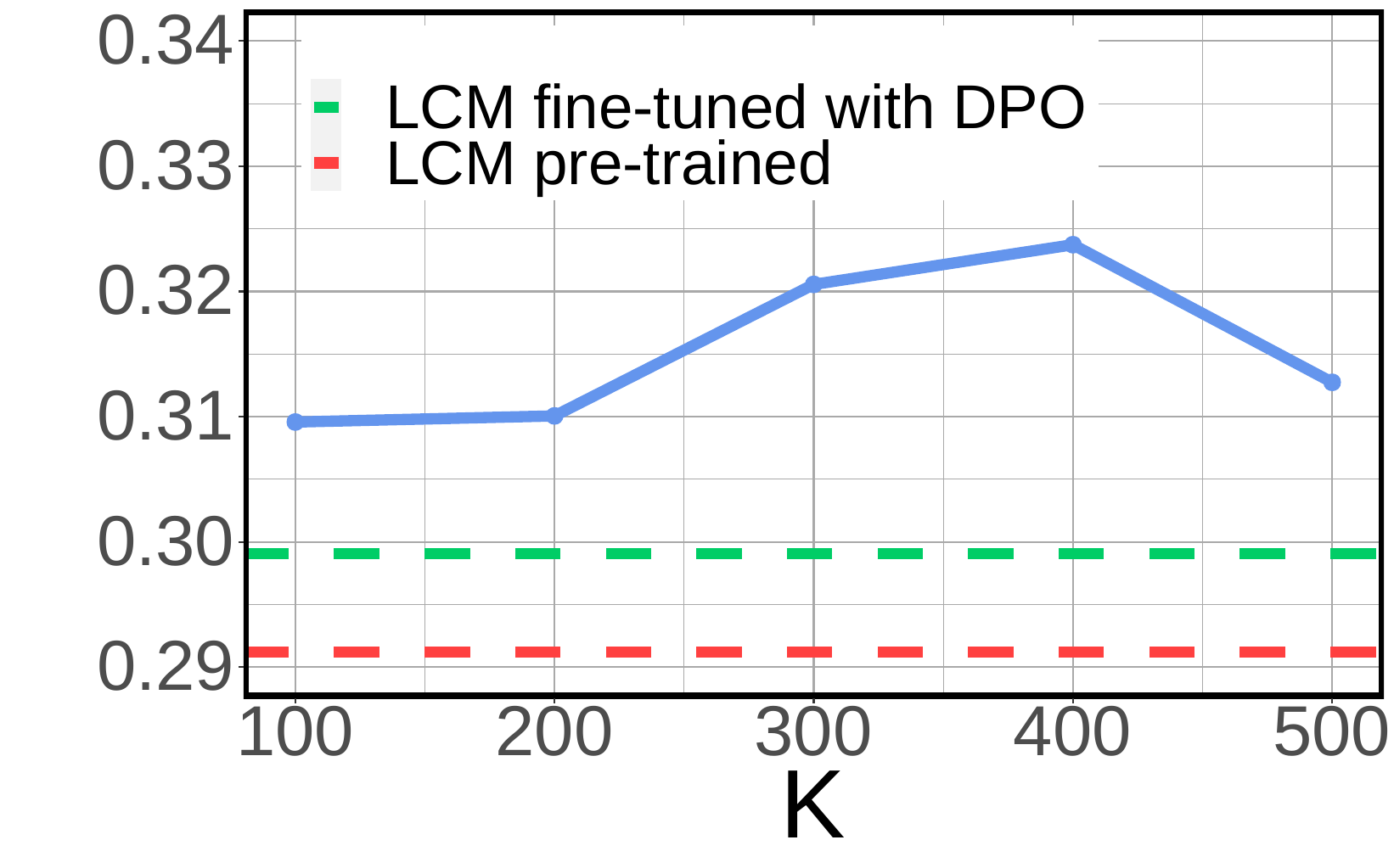}  
\vspace{-0.6cm}
  \caption{Varying $K$ for Curriculum-DPO.}
  \label{fig:sub-K}
\end{subfigure}
\begin{subfigure}[t]{.245\textwidth}
  \centering
  \includegraphics[width=.99\linewidth]{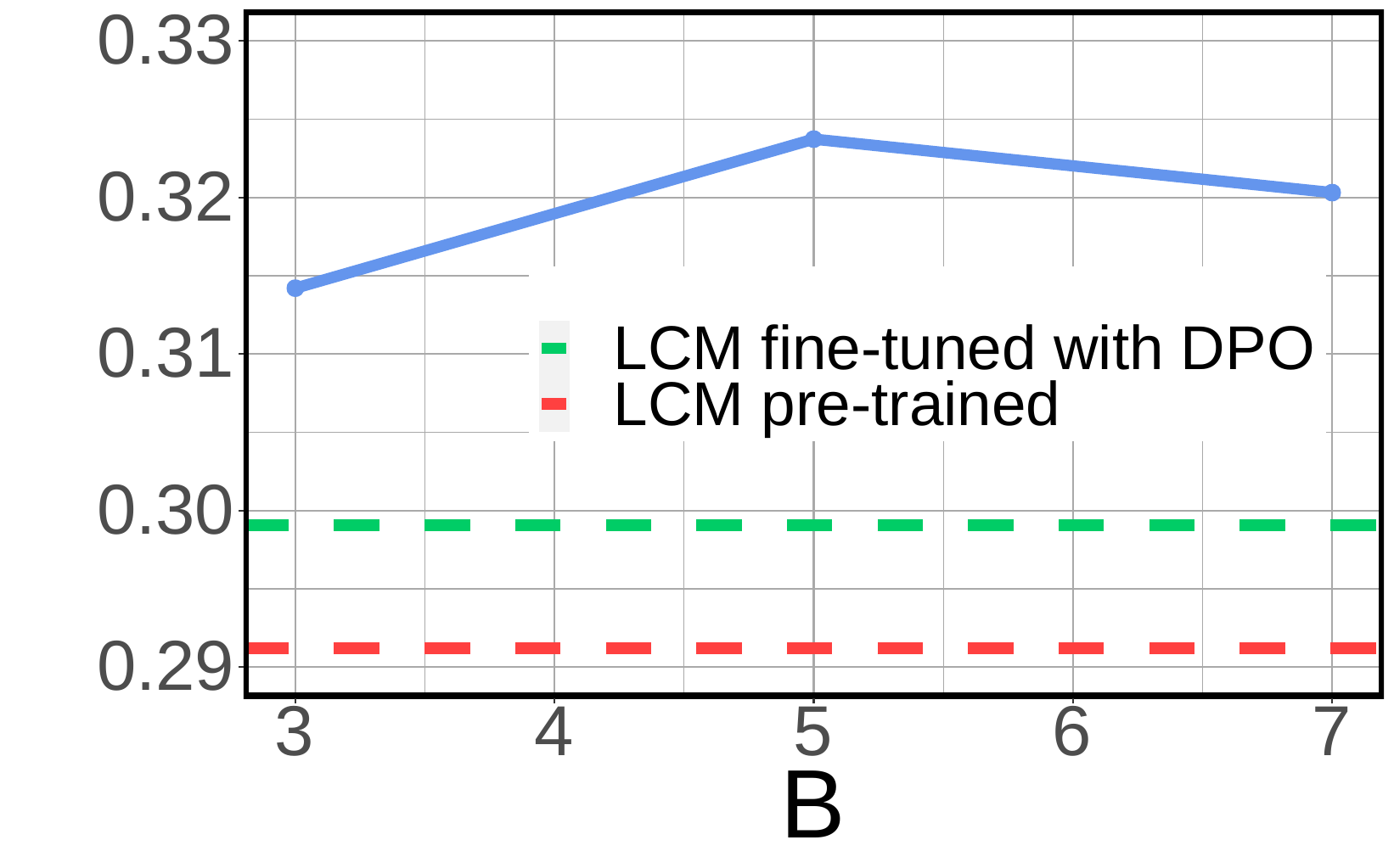} 
  \vspace{-0.6cm}
  \caption{Varying $B$ for Curriculum-DPO.}
  \label{fig:sub-B}
\end{subfigure}
\begin{subfigure}[t]{.245\textwidth}
  \centering
\includegraphics[width=.99\textwidth]{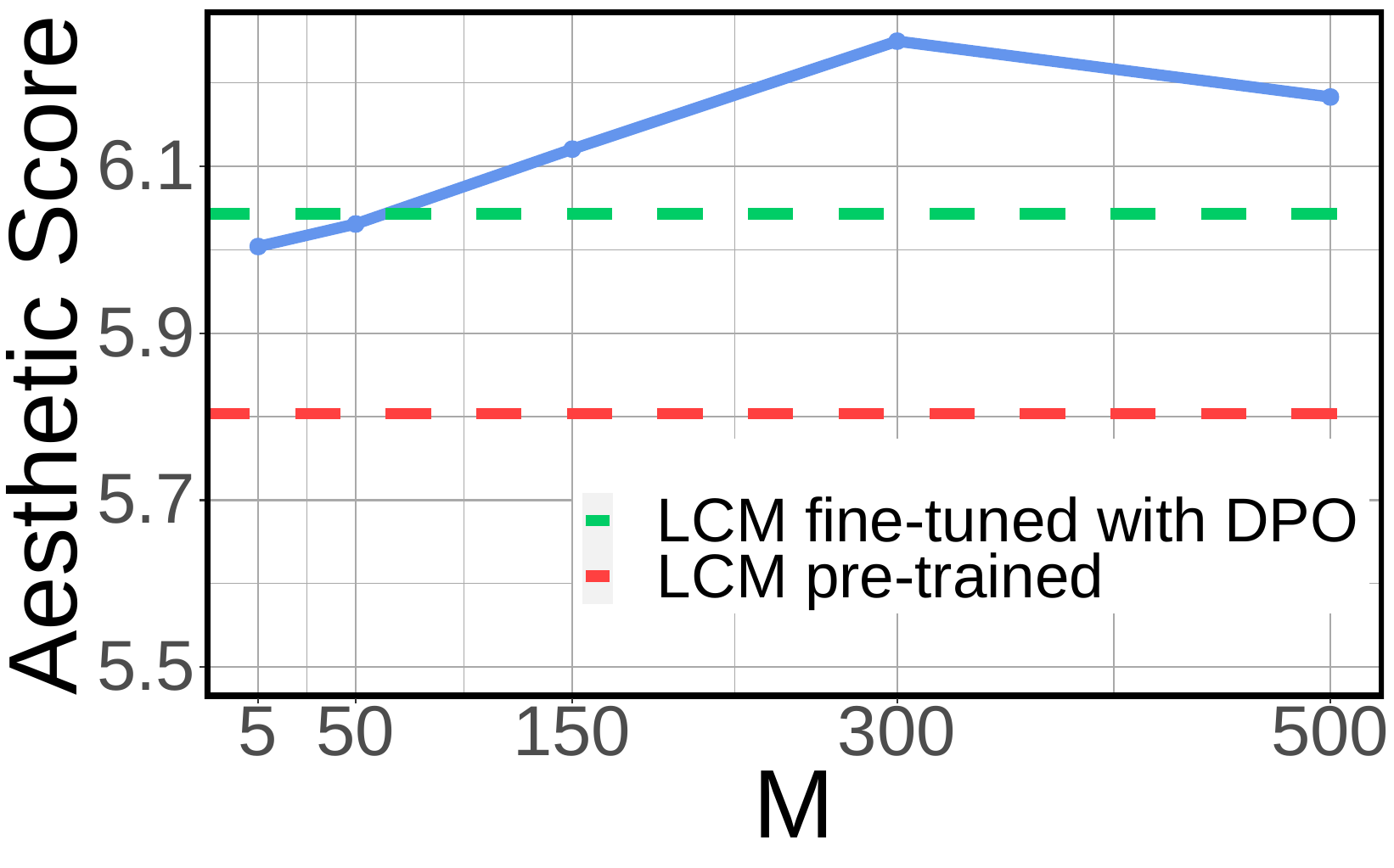}  
\vspace{-0.6cm}
  \caption{Varying $M$ for Curriculum-DPO.}
  \label{fig:sub-M}
\end{subfigure}
\vspace{-0.2cm}
\caption{Ablation results obtained by (a) varying the hyperparameter $\beta$ for Consistency-DPO (in {\color{RoyalBlue}blue}), (b) the number of training iterations per batch $K$ for Curriculum-DPO (in {\color{RoyalBlue}blue}), (c) the number of batches $B$ for Curriculum-DPO (in {\color{RoyalBlue}blue}), and (d) the number of training images per prompt $M$ (in {\color{RoyalBlue}blue}). Fine-tuned LCM models are compared with the pre-trained LCM baseline on the human preference and visual appeal tasks, where scores are given by the HPSv2 reward model and LAION Aesthetics Predictor, respectively. Best viewed in color.}
\vspace{-0.4cm}
\label{fig_ablation}
\end{figure*}

\begin{figure}[t]
  \centering
  \includegraphics[width=1.0\linewidth]{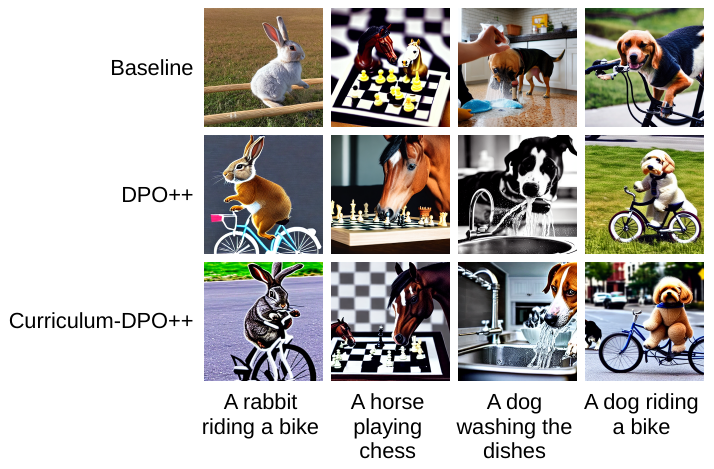}
  \vspace{-0.6cm}
   \caption{Qualitative results on dataset $D_1$, before and after fine-tuning for the text alignment task with our reward-model-free methodology. Best viewed in color.}
   \vspace{-0.2cm}
\label{qualitative_model_free}
\end{figure}

\begin{figure}
    \centering
    \includegraphics[width=0.7\linewidth]{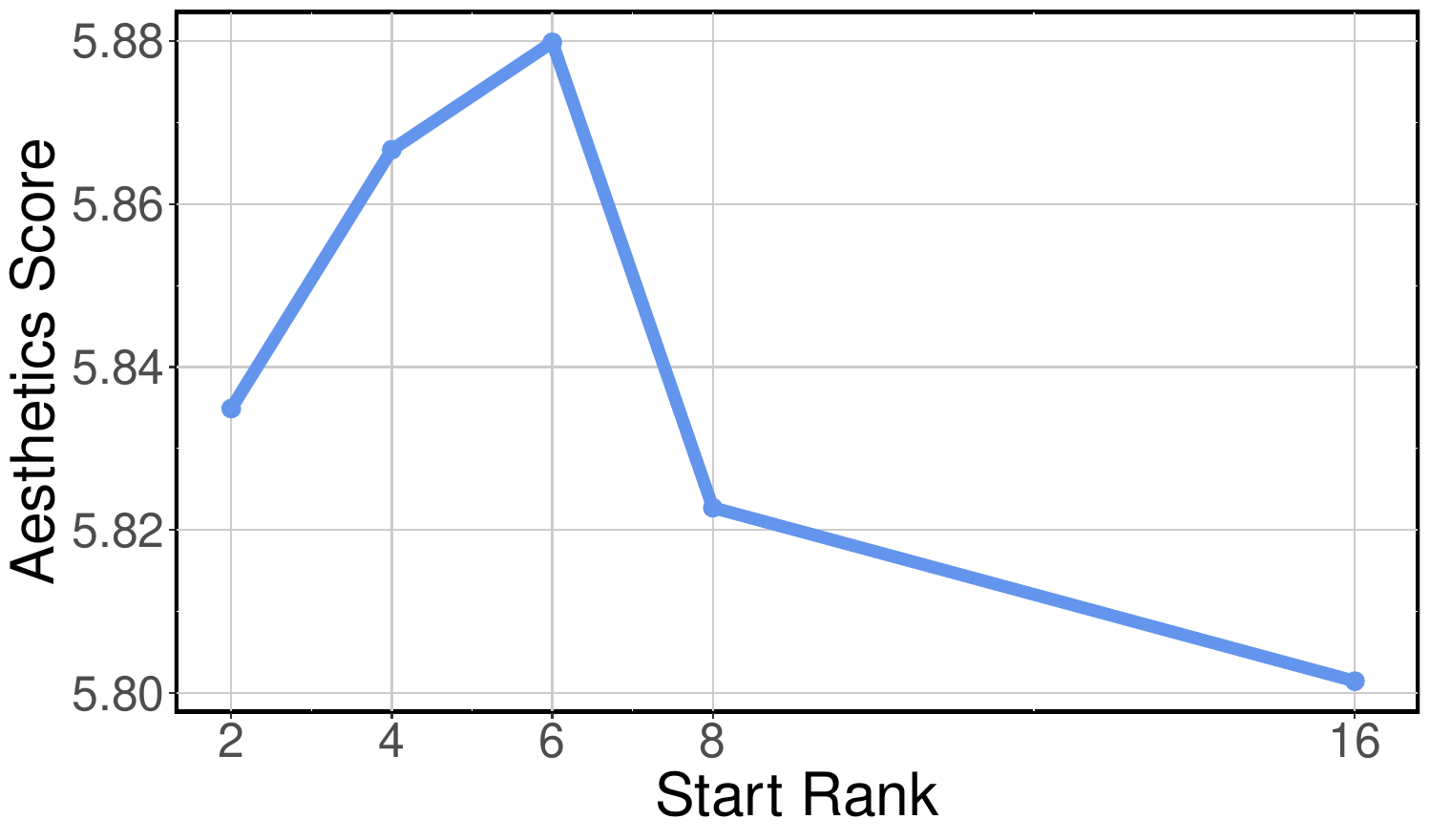}
    \vspace{-0.2cm}
    \caption{Ablation study on the start rank $\mathbf{r}_{\text{start}}$ for the LoRA matrices. The study is conducted for SD on $D_2$ using the LAION Aesthetics reward. For this ablation, the final rank $\mathbf{r}_{\text{end}}$ is fixed to $16$.}
    \label{fig_ablation_r_start}
    \vspace{-0.4cm}
\end{figure}

\noindent
\textbf{Main results.} We present quantitative results for Curriculum-DPO and Curriculum-DPO++ versus their competitors in Table~\ref{tab_strategies}. Our original Curriculum-DPO \cite{Croitoru-CVPR-2025} surpasses both DDPO and DPO in all scenarios. Moreover, Curriculum-DPO++, the version that integrates our new model-level curriculum, yields a considerable performance boost in most cases, the most significant gains being obtained on human preference for SD on all three datasets. We further complement the quantitative evaluation with qualitative results, depicted in Figure~\ref{qualitative_text_align}. The illustrated images are obtained after fine-tuning for text alignment. Although the original Curriculum-DPO does a good job at illustrating animals performing specific activities, our new variant, Curriculum-DPO++, improves the results in several cases. For example, the beetle in the fourth column is generated with a more accurate appearance, and the bear in the seventh column depicts the activity more clearly. Therefore, even in cases where numerical improvements seem limited, such as the text alignment scores for LCM on $D_1$ (see Table~\ref{tab_strategies}), our method demonstrates a distinct perceptual advantage that is clearly visible in the qualitative results. This is further strengthened by the win rates depicted in Figures~\ref{fig_winrates_sd} and~\ref{fig_winrates}. The win rates are computed for each of the three reward models, and are based on how many times each method yields the highest score. More qualitative results are shown in Appendix \ref{supp_qual}.

\noindent
\textbf{Reward-model-free results.}
In Table~\ref{tab_model_free}, we present results for Curriculum-DPO and Curriculum-DPO++ using the reward-model-free strategy to create the preference pairs. The proposed reward-model-free strategy based on masking text embeddings demonstrates substantial performance gains compared with the pre-trained LCM and SD baselines across all experimental settings. We observe that the gains are usually higher for SD. Our model-level curriculum learning strategy yields additional enhancements in all scenarios, except for one (SD on dataset $D_2$). In Figure~\ref{qualitative_model_free}, we showcase some qualitative samples obtained from the fine-tuned SD models on $D_1$. These examples demonstrate that our reward-model-free methods, Curriculum-DPO and Curriculum-DPO++, better align with the prompts, while also being more visually appealing. For instance, the rabbit is riding the bike (in line with the prompt) in our images depicted on the first column, but the rabbit just stays on the fence in the baseline image.

\noindent
\textbf{Ablations.} The original Curriculum-DPO has several hyperparameters. We showcase the impact of each hyperparameter through a series of ablation studies. First, we explore different values of $\beta$, when fine-tuning LCM via Consistency-DPO. Results with $\beta \in \{50,\dots,500\}$ are shown in Figure~\ref{fig:sub-beta}. When $\beta\!\geq\!200$, the fine-tuning outperforms the baseline. In Figure~\ref{fig:sub-K}, we search for the optimal value for the number of training iterations per curriculum batch, $K$, in terms of HPSv2 scores. For all of the explored values between $100$ and $500$, we obtain better results than the baseline. 
Further, in Figure~\ref{fig:sub-B}, we explore how many curriculum batches are needed, considering values between $3$ and $7$. Curriculum-DPO consistently surpasses the baseline, regardless of the number of training batches. Finally, in Figure~\ref{fig:sub-M}, we assess the impact of the number of training images generated per prompt, from $M\!=\!5$ to $M\!=\!500$. For reference, we include the results of DPO and DDPO based on $M\!=\!500$ images per prompt. Notably, Curriculum-DPO is able to obtain comparable performance with DPO and DDPO, while using 10$\times$ less training images. Moreover, Curriculum-DPO outperforms the baseline LCM regardless of the number of training images. In general, we observe that Curriculum-DPO maintains its superiority over DPO under various hyperparameter configurations.

\begin{table}[t]
  
  \centering
  \small{
  \begin{tabular}{ccc}
  \toprule
    {Progressive Rank}     & {Progressive Layers}     & {Aesthetics} \\  
  \midrule
  \textcolor{Red}{\xmark} & \textcolor{Red}{\xmark} &   5.7060 \\
 \textcolor{ForestGreen}{\checkmark} & \textcolor{Red}{\xmark} &   5.7400 \\
 \textcolor{Red}{\xmark} & \textcolor{ForestGreen}{\checkmark} &   5.8015 \\
  \textcolor{ForestGreen}{\checkmark} & \textcolor{ForestGreen}{\checkmark} &   5.8798\\
 \bottomrule
  \end{tabular}
  }
  \vspace{-0.2cm}
  \caption{Ablation study on the key components of the model-level curriculum added on top of Curriculum-DPO. The study is conducted on $D_2$ with the SD fine-tuned via the LAION Aesthetics reward model.}
  \label{tab_ablation_components}
    \vspace{-0.3cm}
\end{table}

We next present additional ablations for the contributions included in Curriculum-DPO++. To determine the optimal value for the start dimension $\mathbf{r}_{\text{start}}$ of the LoRA matrices, we conduct an ablation study using SD on dataset $D_2$, using the LAION Aesthetics reward. To ensure a fair comparison with the original Curriculum-DPO, we do not vary $\mathbf{r}_{\text{end}}$.  Instead, we fixed this parameter to match the baseline configuration exactly. We present the corresponding results in Figure~\ref{fig_ablation_r_start}. Curriculum-DPO reaches an aesthetics score of $5.706$ (see Table \ref{tab_strategies}), while Curriculum-DPO++ is consistently above $5.8$ (see Figure~\ref{fig_ablation_r_start}), indicating that model-level curriculum is useful, no matter how $\mathbf{r}_{\text{start}}$ is set. 

The model-level curriculum progressively increases both the rank of the LoRA matrices and the number of layers that are being updated via LoRA. In Table~\ref{tab_ablation_components}, we showcase the impact of each component of the model-level curriculum, both in isolation as well as in the combined configuration. Two primary conclusions can be drawn from these results. First, we observe that every variation of our method consistently surpasses the baseline, confirming the robustness of our model-level curriculum across all tested scenarios. Second, the results indicate that the two components are effectively orthogonal. Consequently, when combined, they produce a substantial performance boost that exceeds the gains of either component individually.

\vspace{-0.2cm}
\section{Conclusion}
\label{sec: conclusion}
\vspace{-0.1cm}

Building on top of our previous work introducing data-level curriculum into DPO, we proposed an enhanced framework, deemed Curriculum-DPO++, that gracefully integrates data-level and model-level curricula. Curriculum-DPO++ increases learning capacity at the same pace with data complexity, enabling better generalization. Through comprehensive experiments on three datasets and three downstream tasks, we demonstrated consistent performance gains due to our model-level progressive training. Moreover, our second contribution, a reward-model-free extension of Curriculum-DPO++, which is based on text embedding masking, proved to be a viable fine-tuning solution, showing improvements in almost all text alignment and human preference scenarios. In general, Curriculum-DPO++ demonstrated both quantitative and qualitative enhancements, producing models that are more inclined to follow the input prompt and generate images that are more visually appealing. We conducted ablation experiments to attest that each contribution is empirically relevant and to determine the robustness to hyperparameter variations. Our ablation studies showed that Curriculum-DPO and Curriculum-DPO++ are robust to suboptimal hyperparameter settings, obtaining consistent performance gains over the corresponding baselines for a broad range of hyperparameter configurations.

In future work, we intend to apply Curriculum-DPO++ beyond image generation, e.g.~to fine-tune Large Language Models. In natural language processing, curricula could explicitly be defined for textual data, taking into account prompt complexity or reasoning depth.

\vspace{-0.15cm}
\ifCLASSOPTIONcompsoc
  \section*{Acknowledgments}
\else
  \section*{Acknowledgment}
\fi
\vspace{-0.1cm}
This work was supported by a grant of the Ministry of Research, Innovation and Digitization, CNCS -
UEFISCDI, project number PN-IV-P1-PCE-2023-0354, within PNCDI IV.

\vspace{-0.2cm}
{\small
\bibliographystyle{ieeetr}
\bibliography{shortref-nopages}
}

\clearpage

\section{Appendix}

\subsection{Detailed Preliminaries}
\label{supp_prelim}

\textbf{Diffusion models.}
Diffusion models \cite{Croitoru-TPAMI-2023, ho-NeurIPS-2020, karras-CVPR-2020,kingma-NeurIPS-2021, rombach-CVPR-2022, sohl-icml-2015,  song-ICLR-2021, song-NeurIPS-2019,  vahdat-NeurIPS-2021} are a class of generative models trained to reverse a process that progressively inserts Gaussian noise across $T$ steps into the original data samples, transforming them into standard Gaussian noise. Formally, the forward process defined by:
\begin{equation}
\label{eq_forward_process}
        x_t = \alpha_t x_0 + \sigma_t \epsilon, \epsilon \sim \mathcal{N}(0, \mathbf{I})
\end{equation}
transforms the original samples $x_0 \sim p(x_0)$ into noisy versions $x_t$, following the noise schedule implied by the time-dependent predefined functions $(\alpha_t)_{t=1}^{T}$ and $(\sigma_t)_{t=1}^{T}$. The model is trained to estimate the noise $\epsilon$ added in the forward step defined in Eq.~\eqref{eq_forward_process}, by minimizing the following objective:
\begin{equation}
    \mathcal{L}_{\text{simple}} = \mathbb{E}_{t \sim \mathcal{U}(1, T), \epsilon \sim \mathcal{N}(0, \mathbf{I}), x_0 \sim p(x_0)} \!\left\lVert \epsilon_t-\epsilon_\theta(x_t,t)\right\rVert^{2}.
\end{equation}
The generation process involves denoising, starting from a sample of standard Gaussian noise, denoted as \( x_T \sim \mathcal{N}(0, \mathbf{I}) \). It then follows the transitions outlined in Eq.~\eqref{eq_reverse_process} to produce novel samples:
\begin{equation}
    \label{eq_reverse_process}
    p_\theta(x_{t-1}|x_t) = \mathcal{N}\left(x_{t-1}; \mu_\theta(x_t, t), \sigma_{t|t-1}^2\frac{\sigma_{t-1}^2}{\sigma_t^2}\right), 
\end{equation}
with $\mu_\theta(x_t, t) = \frac{1}{\alpha_{t|t-1}}\left(x_t - \frac{\epsilon_\theta(x_t, t)\sigma_{t|t-1}^2}{\sigma_t} \right)$,
where $\sigma_{t|t-1}^2=\sigma_t^2-\alpha_{t|t-1}^2 \sigma_{t-1}^2$ and $\alpha_{t|t-1}=\frac{\alpha_t}{\alpha_{t-1}}$.

The forward process can also be defined in a continuous time manner \cite{song-ICLR-2021}, as a stochastic differential equation (SDE):
\begin{equation}
    \label{eq_forward_process_sde}
    \mathrm{d}x_t = f(t)x_t\mathrm{d}t + g(t)\mathrm{d}\omega_t, t \in [0, T],
\end{equation}
where, given the notations from Eq.~\eqref{eq_forward_process}, we can write $f(t) = \frac{\mathrm{d} \log{\alpha(t)}}{\mathrm{d}t}$ and $g^2(t) = \frac{\mathrm{d} \sigma^2(t)}{\mathrm{d}t} - 2 \frac{\mathrm{d} \log{\alpha(t)}}{\mathrm{d}t} \cdot \sigma^2(t)$, and $\omega_t$ is the standard Brownian motion.

Furthermore, the diffusion process described by the SDE from Eq.~\eqref{eq_forward_process_sde} can be reversed by another diffusion process given by a reverse-time SDE \cite{Andreson-SPA-1982, song-ICLR-2021}. In addition, Song \etal~\cite{song-ICLR-2021} showed that the reverse SDE has a corresponding ordinary differential equation (ODE), called Probability flow-ODE (PF-ODE), with the following form:
\begin{equation}
    \label{eq_ODE_reverse}
    \mathrm{d}x_t = f(t)x_t\mathrm{d}t + \frac{g^2(t)}{2 \sigma(t)}\epsilon_{\theta}(x_t, t).
\end{equation}

\begin{algorithm*}[!t]
\caption{Curriculum-DPO++ (for diffusion models)}
\label{alg:method_diffusion}
\KwIn{$\{(x_{0, i}, c)\}_{i=1}^M$ - the training samples, $r_\varphi(x_0, c)$ - the reward model which can be conditioned on $c$, $B$ - the number of batches for splitting the set of pairs, $\alpha_t, \sigma_t$ - the parameters of the noise schedule, $T$ - the last diffusion time step, $\beta$ - DPO hyperparameter to control the divergence from the initial pre-trained state, $\sigma$ - the sigmoid function, $\eta$ - the learning rate, $\{H_k\}_{k=1}^B$ - the number of training iterations after including the $k$-th batch, $\{G_k\}_{k=1}^B$\ - the trainable layers after including the $k$-th batch, $\phi$ - the pre-trained weights of the reference model, 
$\mathbf{r}_{\text{start}}$ - the rank of the LoRA adaptation matrices at the start of the training, $\mathbf{r}_{\text{end}}$ - the rank of the LoRA adaptation matrices at the end of the training, $\delta$ - the rank update rate.}
\KwOut{
$\theta$ - the trained weights of the generative model.
}
$\hat{X} \leftarrow \{(x_{0,i}, c)|r_\varphi(x_{0,i}, c) \leq r_\varphi(x_{0,i-1},  c), i=\{2,3,...,M\}\};$ \textcolor{commentGreen}{$\lhd$ sort the samples in descending order of the rewards}\\
$S \leftarrow \left\{(x_{0,i}, x_{0, j}, c)| i,j \in \{1, \dots ,M\}; i<j; x_{0,i}, x_{0, j} \in \hat{X},  r_\varphi(x_{0,i}, c) > r_\varphi(x_{0,j}^l, c)  \right\}$; \textcolor{commentGreen}{$\lhd$ create pairs of examples using the order from $\hat{X}$}\\
$\mathrm{L}_k \leftarrow \left\{\frac{(M-1) \cdot(B-k)}{B}\right\}_{k=1}^B$; \textcolor{commentGreen}{$\lhd$ the minimum preference limits of the batches}\\
$\mathrm{R}_k \leftarrow \left\{\frac{(M-1)\cdot(B-(k-1))}{B}\right\}_{k=1}^B$;  \textcolor{commentGreen}{$\lhd$ the maximum preference limits of the batches}\\
$S_k\leftarrow \left\{(x_{0}^w, x_{0}^l, c) | (x_{0}^w, x_{0}^l) = (x_{0,i}, x_{0, j}); \mathrm{L}_k   < j-i \leq \mathrm{R}_k  ; (x_{0,i}, x_{0, j}, c) \in S \right\}_{k = 1}^B$;  $\lhd$  \textcolor{commentGreen}{the batches of increasingly difficult pairs}\\

$\mathbf{r} \leftarrow \left\{\mathbf{r}_k|k \in \{1, \dots , B\}, \mathbf{r}_1 = \mathbf{r}_{\text{start}}, \mathbf{r}_{k+1} = \min\left(\mathbf{r}_{\text{end}}, \mathbf{r}_{k} \cdot \delta\right)\right\}$; \textcolor{commentGreen}{$\lhd$ the ranks of the low-rank matrices in LoRA}\\

$\text{LoRA}_k = \left\{\left(A^{\mathbf{r}_k}_i, C^{\mathbf{r}_k}_i\right) | i \in G_k,A^{\mathbf{r}_k}_i \sim \mathcal{N}(0,\sigma), C^{\mathbf{r}_k}_i = \mathbf{0}_{m\times \mathbf{r}_k} \right\}_{k=1}^B$; \textcolor{commentGreen}{$\lhd$ initialization of LoRA matrices} \\

$P \leftarrow \emptyset$;  \textcolor{commentGreen}{$\lhd$ current training set} \\

$G \leftarrow \emptyset$; \textcolor{commentGreen}{$\lhd$ current trainable layers}\\
\ForEach{$k \in \{1, \dots, B\}$}
{
    $P \leftarrow P \cup S_k$; \textcolor{commentGreen}{$\lhd$ include a new batch in the training}\\
    
    $G^{\text{new}} \leftarrow \text{LoRA}_k$; \textcolor{commentGreen}{$\lhd$ include new set of trainable layers}\\
    \ForEach{ $\left(A^{\mathbf{r}_k}_i, C^{\mathbf{r}_k}_i\right) \in G^{\textnormal{new}}$}
    {
        \If{$i < |G|$}{ \textcolor{commentGreen}{// transfer the previously learned weights to the new matrices} \\
        $A^{\mathbf{r}_{k-1}}_i, C^{\mathbf{r}_{k-1}}_i \leftarrow G_i $; \\
        $A^{\mathbf{r_{k}}}_i\left[:\mathbf{r}_{k-1}, :\right] \leftarrow A^{\mathbf{r}_{k-1}}_i$; \\

        $C^{\mathbf{r_{k}}}_i\left[:,:\mathbf{r}_{k-1}\right] \leftarrow C^{\mathbf{r}_{k-1}}_i$; \\
        
        }

    }
    $G \leftarrow G^{\text{new}}$;\\
    \ForEach{$i \in \{1, \dots, H_k\}$}
    {
    $(x_0^w, x_0^l, c) \sim \mathcal{U}(P)$; 
    $t \sim \mathcal{U}\{1, \dots, T\}$; $\epsilon^w, \epsilon^l \sim \mathcal{N}(0, \mathbf{I})$;\\
    $x_{t}^w \leftarrow \alpha_{t} x_0^w + \sigma_{t} \epsilon^w$;  \textcolor{commentGreen}{$\lhd$ forward process} \\
    $x_{t}^l \leftarrow \alpha_{t} x_0^l + \sigma_{t} \epsilon^l$; \textcolor{commentGreen}{$\lhd$ forward process}\\
    
    $\mathcal{L}_{\mathrm{\text{Diff-DPO}}}(\theta, G)\! \leftarrow \! - \Big[\!\log\sigma\Big(\!\!-\!\beta T 
        \Big(\!\!
    \left(\lVert\epsilon^w - \epsilon_{\theta,G}^w(x_t^w, t, c) \rVert_2^2 -
    \lVert\epsilon^w - \epsilon_{\text{ref}}^w(x_t^w, t, c)\rVert_2^2
    \right) \!-\!  
    \left(\lVert\epsilon^l - \epsilon_{\theta,G}^l(x_t^l, t, c)\rVert_2^2 -
    \lVert\epsilon^l - \epsilon_{\text{ref}}^l(x_t^l, t, c)\rVert_2^2
    \right)\!\!
    \Big)\!\! \Big)\! \Big]$; \textcolor{commentGreen} {$\lhd$ DPO loss}\\
    $G \leftarrow G - \eta \frac{\partial \mathcal{L}_\mathrm{\text{Diff-DPO}}}{\partial G}$; \textcolor{commentGreen}{$\lhd$ update the weights}
    }   
}
$\theta \leftarrow \theta + G$; \textcolor{commentGreen}{$\lhd$ store the LoRA weights into the original model}\\
\textbf{return} $\theta$
\end{algorithm*}

\noindent
\textbf{Consistency models.} Consistency models \cite{Luo-arXiv-2023, Song-ICLR-2024,  Song-ICML-2023} are a new class of generative models. These models operate on the idea of training a model to associate each point along a trajectory of the PF-ODE (Eq.~\eqref{eq_ODE_reverse}) to the initial point of the respective trajectory, which corresponds to the denoised sample. Such models can either be trained from scratch or through distillation from a pre-trained diffusion model. In our study, we employ the distillation method, so we next detail this approach.

Given a solution trajectory $\{x_t\}_{t\in \left[\delta, T\right]}$ of the PF-ODE defined in Eq.~\eqref{eq_ODE_reverse}, where $\delta \rightarrow 0$, the training of a consistency model $f_\phi(x_t, t)$ involves enforcing the self-consistency property across this trajectory, such that, $\forall t, t' \in \left[ \delta, T \right]$, the condition $f_\phi(x_t, t) = f_\phi(x_{t'}, t')$ holds. The loss function designed to achieve this self-consistency is described as follows: 
\begin{equation}
\label{app_eq_consistency_distillation}
    \mathcal{L}_{\text{CD}}(\phi) = d(f_\phi(x_{t_{n+1}}, t_{n+1}), f_{\phi^{-}}(\hat{x}_{t_n}^{\theta}, t_n)),
\end{equation}
where $d$ is a distance metric, $n \sim \mathcal{U}(1, N)$, $N$ is the discretization length of the interval $\left[0, T\right]$, $\phi$ are the trainable parameters of the consistency model and $\phi^{-}$ is a running average of $\phi$. The term $\hat{x}_{t_n}^\theta$ represents a one-step denoised version of $x_{t_{n+1}}$, obtained by applying an ODE solver on the PF-ODE. The solver operates using a pre-trained diffusion model, $\epsilon_\theta(x_{t_n}, t_n)$.

\noindent
\textbf{Direct Preference Optimization (DPO).}
Training pipelines based on Reinforcement Learning with Human Feedback (RLHF) \cite{Ziegler-arXiv-2020} have been highly successful in aligning Large Language Models to human preferences. These pipelines feature an initial phase where a reward model is trained using examples ranked by humans, followed by a reinforcement learning phase where the policy model is fine-tuned to align with the learned reward model. In this context, Rafailov~\etal~\cite{Rafailov-NeurIPS-2023} introduced DPO as an alternative to the previous pipeline, which bypasses the training of the reward model and directly optimizes the policy model using the ranked examples.

\begin{figure*}[t]
  \centering
  \includegraphics[width=1.\linewidth]{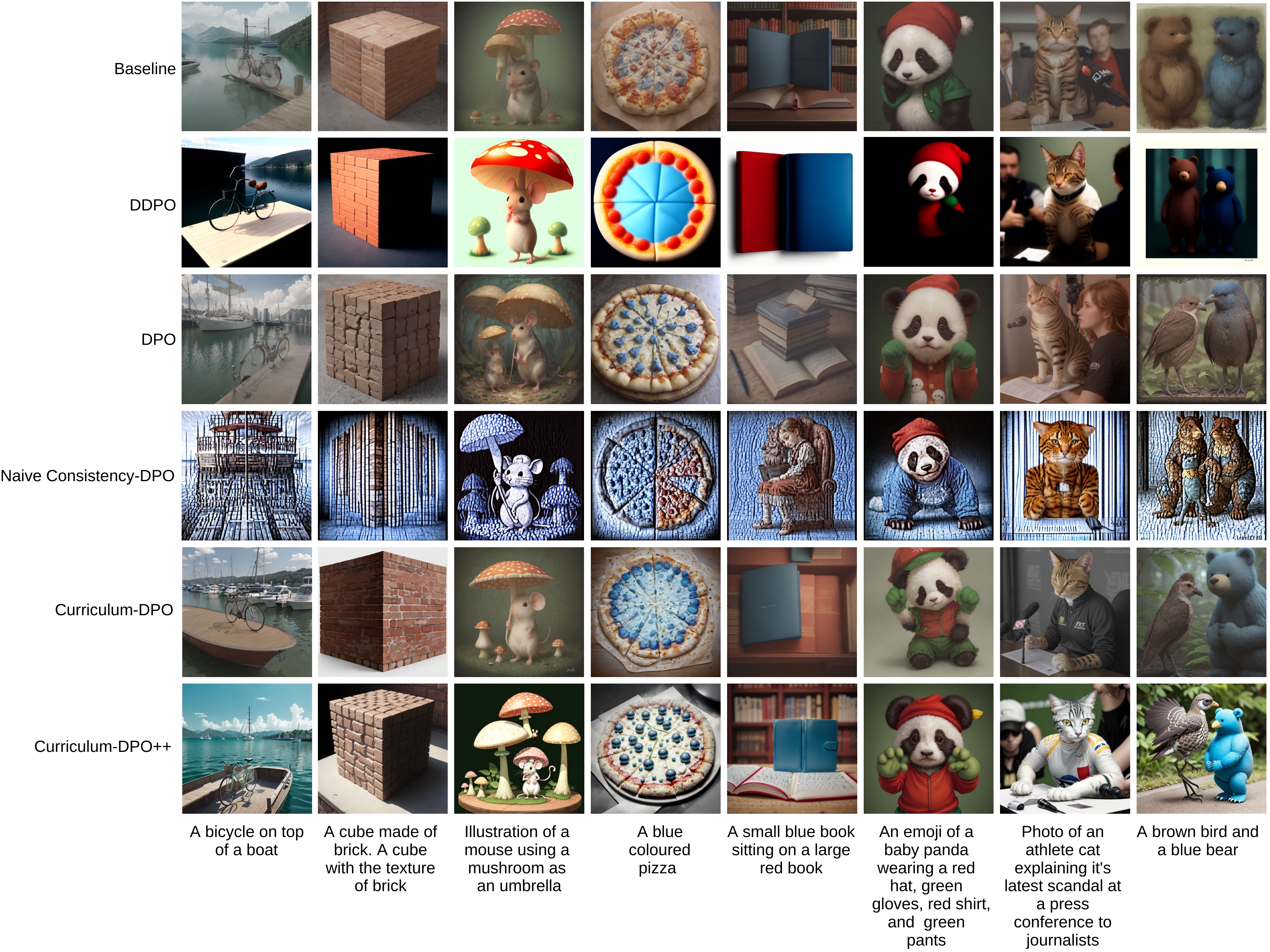}
  \vspace{-0.6cm}
   \caption{Qualitative results before and after fine-tuning for the text alignment task on DrawBench. The fine-tuning methods are: DDPO, DPO, Naive Consistency-DPO, Curriculum-DPO, and Curriculum-DPO++. Best viewed in color.}
   \vspace{-0.3cm}
   \label{drawbnech_qualitative_text_align}
\end{figure*}

The training dataset contains triplets of the form $(c, x_0^w, x_0^l)$, where $x_0^w$ denotes the favored (winning) sample, $x_0^l$ the unfavored (losing) one, and $c$ is a condition used to generate both samples. RLHF trains a reward model by maximizing the likelihood $p(x_0^w \prec x_0^l|c)$\footnote{$a \prec b$ denotes that $a$ precedes $b$ in the ranking implied by the reward model.}, which, under the Bradley-Terry (BT) model, has the following form:
\begin{equation}
    p_{\text{BT}}(x_0^w \prec x_0^l|c) = \sigma(r_\varphi(x_0^w, c) - r_\varphi(x_0^l, c)),
\end{equation}
where $\sigma$ denotes the sigmoid function and $r_\varphi$ is the reward model parameterized by the trainable parameters $\varphi$. The training objective for the reward model is the negative log-likelihood:
\begin{equation}
    \label{eq_neg_log}
    \mathcal{L}_{\text{BT}} = - \mathbb{E}_{x_0^w, x_0^l, c}\left[\log{\sigma(r_\varphi(x_0^w, c) - r_\varphi(x_0^l, c))}\right].
\end{equation}
After training the reward model $r_\varphi(x_0, c)$, RLHF optimizes a conditional generative model $p_\theta(x_0|c)$ to maximize the reward $r_\varphi(x_0, c)$ and, at the same time, controls the deviance from a reference model $p_{\text{ref}}(x_0, c)$ through a Kullback–Leibler (KL) divergence term:
\begin{equation}
    \label{eq_policy_opt}
    \max_{\theta} \mathbb{E}_{c, x_0 \sim p_\theta(x_0|c)}\left[ r_\varphi(x_0, c)\!-\!\beta\!\cdot\!\text{KL}(p_\theta(x_0|c), p_{\text{ref}}(x_0| c))\right]\!,
\end{equation}
where $\beta$ controls the importance of the divergence term.

\begin{figure*}[t]  
  \centering
  \includegraphics[width=1.\linewidth]{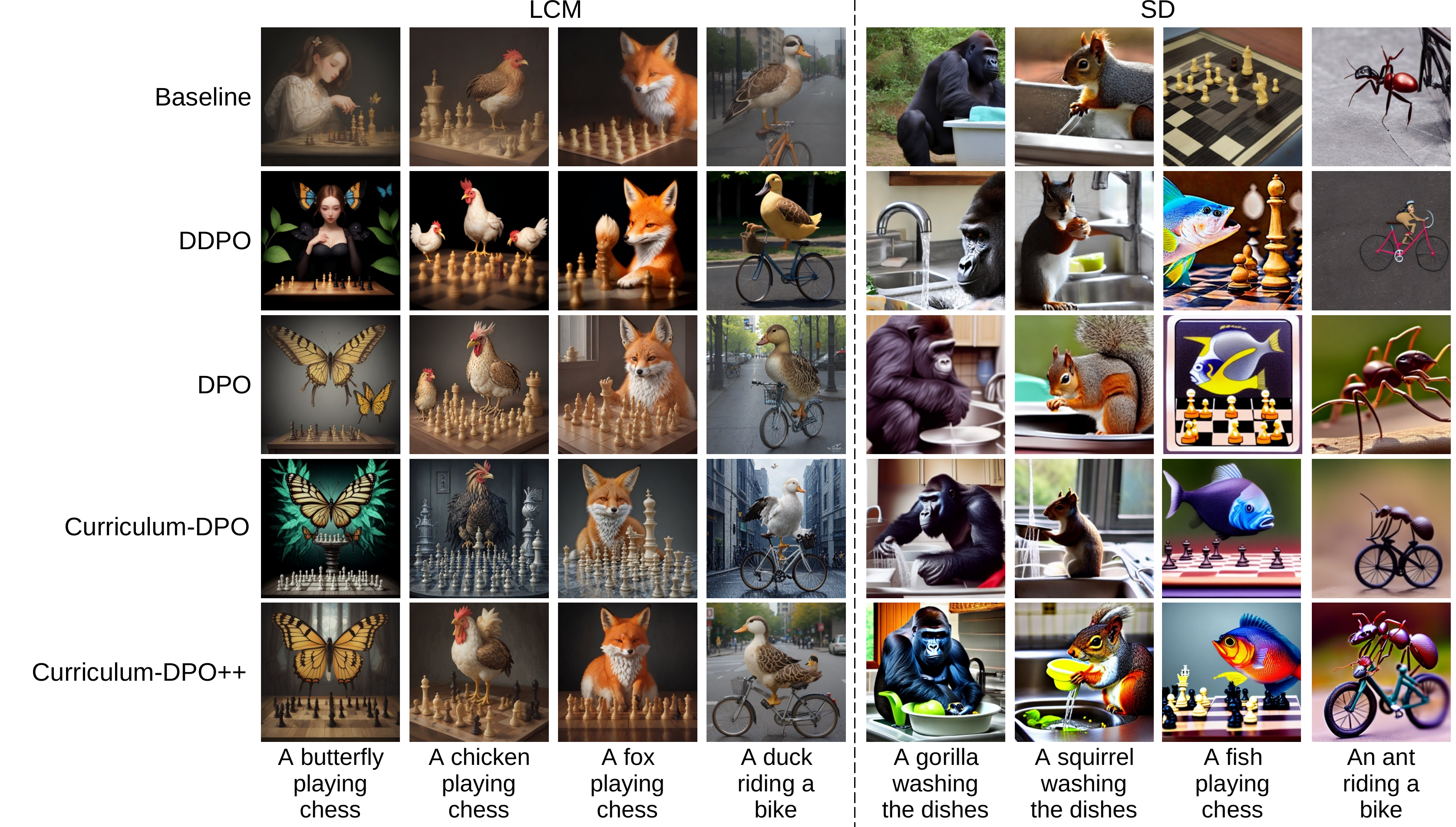}
  \vspace{-0.5cm}
  \caption{Qualitative results after fine-tuning with HPSv2 as the reward model (human preference). The fine-tuning alternatives are: DDPO, DPO, Curriculum-DPO, and Curriculum-DPO++. Best viewed in color.}
  \label{qualitative_human_preference}
  \vspace{-0.2cm}
\end{figure*}

\begin{figure*}[t] 
  \centering
  \includegraphics[width=1.\linewidth]{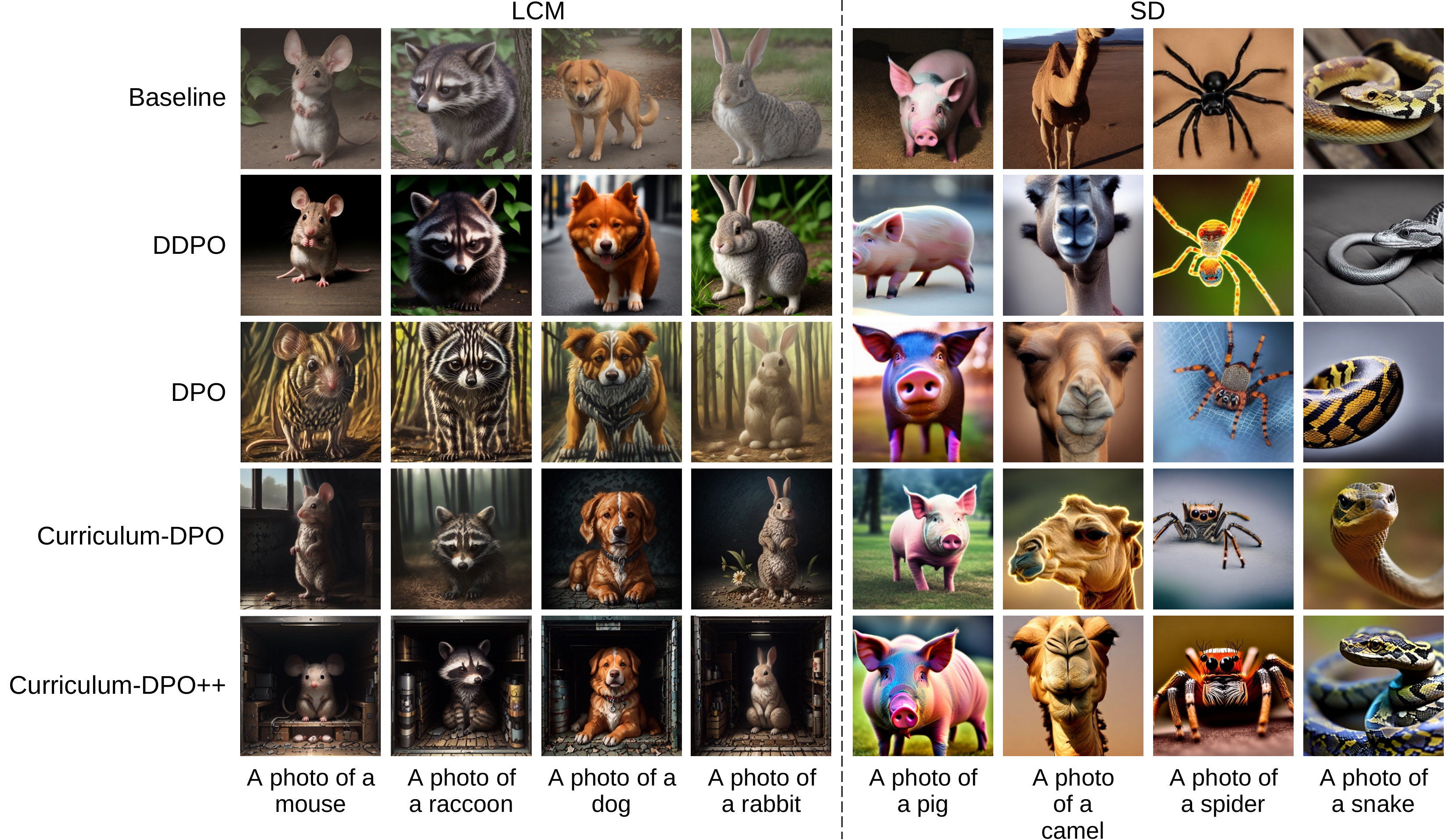}
  \vspace{-0.5cm}
  \caption{Qualitative results after fine-tuning with the LAION Aesthetics Predictor as the reward model. The fine-tuning alternatives are: DDPO, DPO Curriculum-DPO, and Curriculum-DPO++. Best viewed in color.}
  \label{qualitative_aesthetics}
  \vspace{-0.2cm}
\end{figure*}

To derive the DPO objective, Rafailov~\etal~\cite{Rafailov-NeurIPS-2023} write the optimal policy model $p_\theta^*$ of Eq.~\eqref{eq_policy_opt} as a function of the reward and the reference model, as shown in prior works~\cite{Peng-arXiv-2019,Peters-ICML-2007}:
\begin{equation}
    \label{eq_optimal_policy}
    p_{\theta}^{*}(x_0|c) = \frac{p_{\text{ref}}(x_0|c) \cdot \exp\left(\frac{r(x_0, c)}{\beta}\right)}{Z(c)},
\end{equation}
where $Z(c) = \sum_{x_0}{p_{\text{ref}}(x_0|c) \cdot \exp\left(\frac{r(x_0, c)}{\beta}\right)}$ is a normalization constant. Further, from Eq.~\eqref{eq_optimal_policy}, Rafailov~\etal~\cite{Rafailov-NeurIPS-2023} rewrite the reward as:
\begin{equation}
    \label{eq_reward_function}
    r(x_0, c) = \beta\left( \log \frac{p_{\theta}^{*}(x_0|c)}{p_{\text{ref}}(x_0|c)} + \log Z(c)\right).
\end{equation}
Finally, the DPO objective is obtained after replacing the reward in Eq.~\eqref{eq_neg_log} with the form from Eq.~\eqref{eq_reward_function}:
\begin{equation}
    \label{app_eq_dpo}
        \begin{split}
        \mathcal{L}_{\text{DPO}}(\theta)\!\!=\!\!-\mathbb{E}_{x_0^w\!, x_0^l, c} \!\left[\!\log\!\sigma\!\left(\!\!\beta\!\! 
        \left(\!\!\log\!{\frac{p_\theta(x_0^w|c)}{p_{\text{ref}}(x_0^w|c)}}\!-\!\log\!{\frac{p_\theta(x_0^l|c)}{p_{\text{ref}}(x_0^l|c)}}\!\!\right)\!\!\!\right)\!\!\right]\!\!.
    \end{split} 
\end{equation}
To grasp the intuition behind $\mathcal{L}_{\text{DPO}}$, we can analyze  its gradient with respect to $\theta$:
\begin{equation}
    \label{app_eq_grad_dpo}
    \begin{split}
            \frac{\partial \mathcal{L}_{\text{DPO}}(\theta)}{\partial\theta}\!=\!&-\beta \cdot \mathbb{E}_{x_0^w, x_0^l, c}\Bigg[\sigma\!\left( \hat{r}_\theta(x_0^l, c) - \hat{r}_\theta(x_0^w, c)\right)\!\cdot\!\\&\left( \frac{\partial\log{p_\theta(x_0^w|c)}}{\partial\theta} - \frac{\partial\log{p_\theta(x_0^l|c)}}{\partial\theta} \right)\!\Bigg],
    \end{split}
\end{equation}
with $\hat{r}_{\theta}(x_0,c)=\beta\cdot\log{\frac{p_\theta(x_0|c)}{p_{\text{ref}}(x_0|c)}}$. By analyzing Eq.~\eqref{app_eq_grad_dpo}, as discussed in \cite{Rafailov-NeurIPS-2023}, it is evident that the DPO objective enhances the likelihood of favored examples, while diminishing it for the unfavored ones. The magnitude of the update is proportional to the error in $\hat{r}_\theta$. Here, the term ``error'' refers to the degree to which $\hat{r}_\theta$ incorrectly prioritizes the sample $x_0^l$.

\subsection{Curriculum-DPO++ for Diffusion Models}
\label{supp_alg}
We formally present the application of Curriculum-DPO++ to diffusion models in Algorithm~\ref{alg:method_diffusion}. The initial steps 1-19, which outline the curriculum strategy, are identical with those used in the implementation for consistency models described in Algorithm~\ref{alg:method}. Steps 20-24 are changed to include the forward process for the preferred and less preferred samples, along with the Diffusion-DPO loss defined in Eq.~\eqref{eq_diffusion_dpo}.

\subsection{More Qualitative Results}
\label{supp_qual}

In Figure~\ref{drawbnech_qualitative_text_align}, we show qualitative results when fine-tuning the model for text alignment on $D_2$ (DrawBench). In addition to images generated by baselines, DPO, DDPO, Curriculum-DPO and Curriculum-DPO++, we also include images generated by a naive implementation of Consistency-DPO. This implementation refers to the most direct adaptation of Diffusion-DPO to consistency models. More precisely, we substitute the noise estimation in the Diffusion-DPO objective with the consistency distillation loss used in consistency models. However, applying this modification directly breaks the consistency property required by these models and leads to poor results, as illustrated in the 4th row of Figure~\ref{drawbnech_qualitative_text_align}.

In Figures~\ref{qualitative_human_preference} and \ref{qualitative_aesthetics}, we present qualitative results after fine-tuning the models with HPSv2 and LAION Aesthetics Predictor as reward models on $D_1$, respectively. Fine-tuning for human preference (Figure~\ref{qualitative_human_preference}) generally results in generating images with more details for the LCM model. Curriculum-DPO and Curriculum-DPO++ produce better aesthetics for both foreground objects and the background. For the SD model, Curriculum-DPO and Curriculum-DPO++ exhibit a better alignment with the text prompt, our methods being the only ones able to generate the ant on a bike, as displayed in the last column. Moreover, there are cases where Curriculum-DPO++ synthesizes the animals and the activities more clearly than Curriculum-DPO, as shown in the ``squirrel washing the dishes'' example. 
Fine-tuning to improve the visual appeal (Figure \ref{qualitative_aesthetics}) translates into better aesthetics for the animals, as expected. However, Curriculum-DPO and Curriculum-DPO++ return several examples that look better than the rest, \eg~the camel in the sixth column and the dog in the third column.

\subsection{Samples with Masked Prompt Embeddings}
\label{supp_masked_samples}

\begin{figure*}[!t]  
  \centering
  \includegraphics[width=0.85\linewidth]{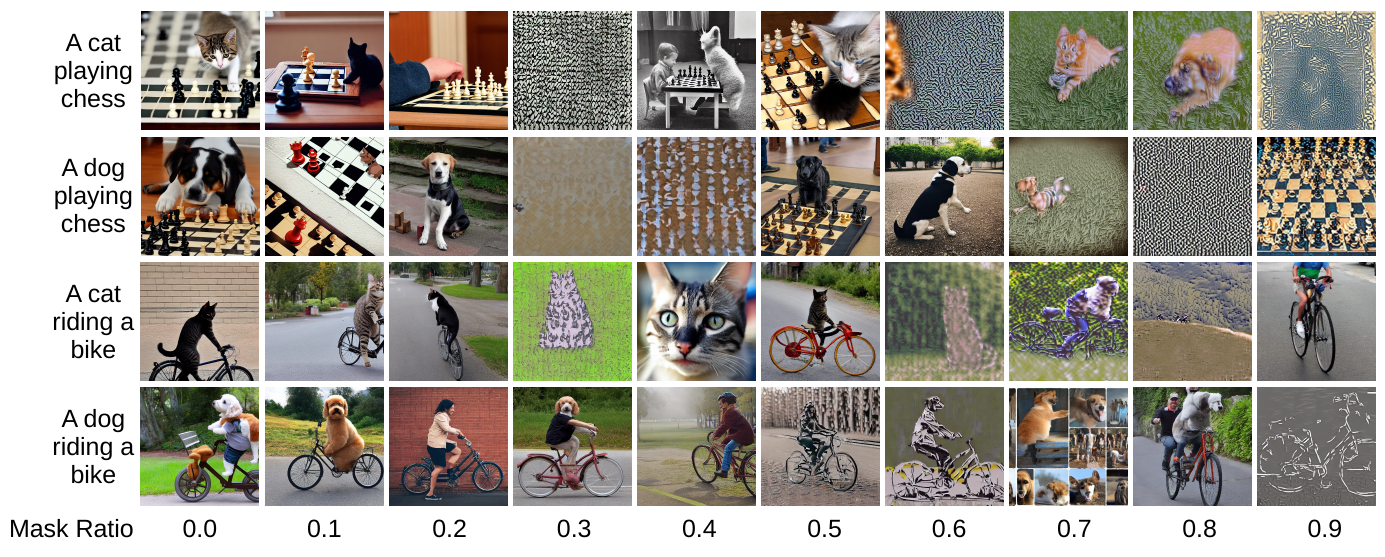}
  \vspace{-0.3cm}
  \caption{Samples generated with SDv1.5 that have the prompt embedding masked with a ratio between $0.0$ and $0.9$. Masking more of the text embeddings leads to lower quality images, enabling a reward-model-free curriculum. Best viewed in color.}
  \label{masked_samples}
  \vspace{-0.2cm}
\end{figure*}

In Figure \ref{masked_samples}, we present examples of images generated while masking the corresponding input prompt embeddings. For each prompt, we showcase samples obtained under different masking ratios. As the masking ratio increases, the generated images progressively degrade, exhibiting reduced alignment with the textual description and a noticeable drop in visual quality, ranging from poorly defined structures to complete noise. The illustrated examples confirm that masking the text embeddings enables the generation of preference pairs with controllable differences between winning and losing examples, without having to use a reward model. By controlling the generation of preference pairs via the masking ratio, we thus obtain a reward-model-free data-level curriculum.
\begin{figure}
    \centering
    \includegraphics[width=1\linewidth]{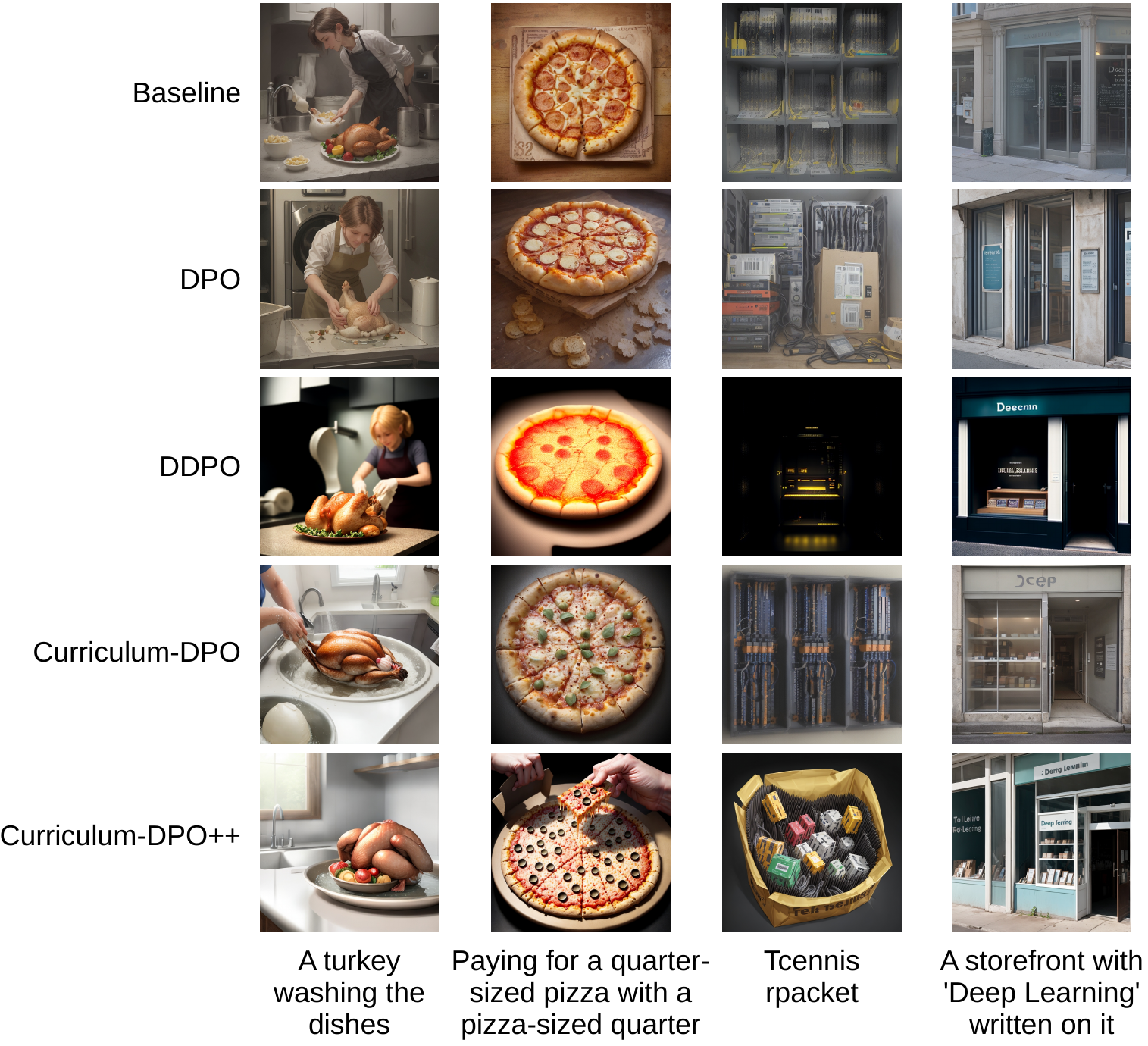}
    \caption{Failure cases of the LCM baseline and fine-tuned variants. In the first column, the models consistently generate a meal to depict the turkey, instead of a bird. In the second column, the models struggle to understand the prompt, they generate a pizza, but they fail to showcase the action of paying for one. The third column demonstrates that the models are not robust to misspellings, and, in the last column, the models fail to generate the required text.}
    \label{fig:failure_cases}
\end{figure}
\subsection{Limitations}
\label{sec: limitations}

One limitation of our framework is the introduction of additional hyperparameters, such as $B$ or $K$. These might require tuning in order to find the optimal values, which involves more computing power. This is a downside of curriculum learning \cite{Soviany-IJCV-2022}, which inherently implies the addition of hyperparameters to control the curriculum learning process. Nevertheless, in the ablation study from Section~\ref{sec: experiments}, we demonstrate that Curriculum-DPO and Curriculum-DPO++ outperform all baselines for multiple hyperparameter combinations. Therefore, suboptimal hyperparameter choices can still improve the generative models.

A limitation of text-to-image generative models (as well as reward models) is the poor ability to disambiguate words in the input prompt. This can be observed especially in the prompt alignment task, where a word with multiple meanings or connotations leads to generating poor results. For example, the prompt ``a turkey washing the dishes'' often results in images of a cooked meal instead of a live bird, as depicted in the first column of Figure~\ref{fig:failure_cases}. In the remaining columns, we show a few other failure cases of the LCM baseline and its fine-tuned variants. In the second column, we show that the models fail to understand complex actions, they simply generate a pizza for a prompt that involves the action of paying for one. The examples included in the third column reveal a lack of robustness to misspellings, and, in the last column, the examples indicate that text generation is only marginally improved. Curriculum learning does not specifically address these limitations of generative and reward models.





\end{document}